\documentclass[letterpaper]{article} 
\usepackage{aaai25}  
\usepackage{times}  
\usepackage{helvet}  
\usepackage{courier}  
\usepackage[hyphens]{url}  
\usepackage{natbib}  
\usepackage{caption} 
\usepackage{algorithm}
\usepackage{algorithmic}
\usepackage{subfigure}
\usepackage[T1]{fontenc}
\usepackage{aecompl}
\usepackage{multirow}
\usepackage{makecell}
\usepackage{booktabs}
\usepackage{amsthm}
\usepackage{amsmath}
\usepackage{xpatch}
\usepackage{amssymb}
\newtheorem{myDef}{Definition}
\usepackage[most]{tcolorbox}

\usepackage{newfloat}
\usepackage{listings}
\DeclareCaptionStyle{ruled}{labelfont=normalfont,labelsep=colon,strut=off} 
\floatstyle{ruled}
\newfloat{listing}{tb}{lst}{}
\floatname{listing}{Listing}
\title{Vision-Based Generic Potential Function for Policy Alignment in Multi-Agent Reinforcement Learning}

\author{
    Hao Ma\textsuperscript{\rm 1,2}\equalcontrib,
    Shijie Wang\textsuperscript{\rm 1,2}\equalcontrib,
    Zhiqiang Pu\textsuperscript{\rm 1,2}\thanks{Corresponding author: zhiqiang.pu@ia.ac.cn.},
    Siyao Zhao\textsuperscript{\rm 1,2},
    Xiaolin Ai\textsuperscript{\rm 2}
}

\affiliations{
    \textsuperscript{\rm 1}School of Artificial Intelligence, University of Chinese Academy of Sciences\\
    \textsuperscript{\rm 2}Institute of Automation, Chinese Academy of Sciences\\
    \{mahao2021, wangshijie2020, zhiqiang.pu, zhaosiyao2023, xiaolin.ai\}@ia.ac.cn
}

\usepackage{bibentry}

\begin{document}

\maketitle

\begin{abstract}
Guiding the policy of multi-agent reinforcement learning to align with human common sense is a difficult problem, largely due to the complexity of modeling common sense as a reward, especially in complex and long-horizon multi-agent tasks. Recent works have shown the effectiveness of reward shaping, such as potential-based rewards, to enhance policy alignment. The existing works, however, primarily rely on experts to design rule-based rewards, which are often labor-intensive and lack a high-level semantic understanding of common sense. To solve this problem, we propose a hierarchical vision-based reward shaping method. At the bottom layer, a visual-language model (VLM) serves as a generic potential function, guiding the policy to align with human common sense through its intrinsic semantic understanding. To help the policy adapts to uncertainty and changes in long-horizon tasks, the top layer features an adaptive skill selection module based on a visual large language model (vLLM). The module uses instructions, video replays, and training records to dynamically select suitable potential function from a pre-designed pool. Besides, our method is theoretically proven to preserve the optimal policy. Extensive experiments conducted in the Google Research Football environment demonstrate that our method not only achieves a higher win rate but also effectively aligns the policy with human common sense.
\end{abstract}

%

\section{Introduction}

Multi-agent reinforcement learning (MARL) has achieved great success in solving complex decision-making problems across lots of domains, including real-time strategy games \cite{vinyals2019grandmaster}, autonomous driving teams \cite{xu2018multi}, robotic systems \cite{yang2004multiagent, gu2023safe}, and sports games \cite{liu2023lazy}. However, MARL faces a very challenging issue: it is difficult to learn policies that conform to the real-world cognition of the domain in complex tasks, especially with sparse rewards \cite{zhu2018human,dossa2019human}.

\begin{figure}[thbp]
 \centering
 \subfigure[]{
    \includegraphics[width=4.05cm]{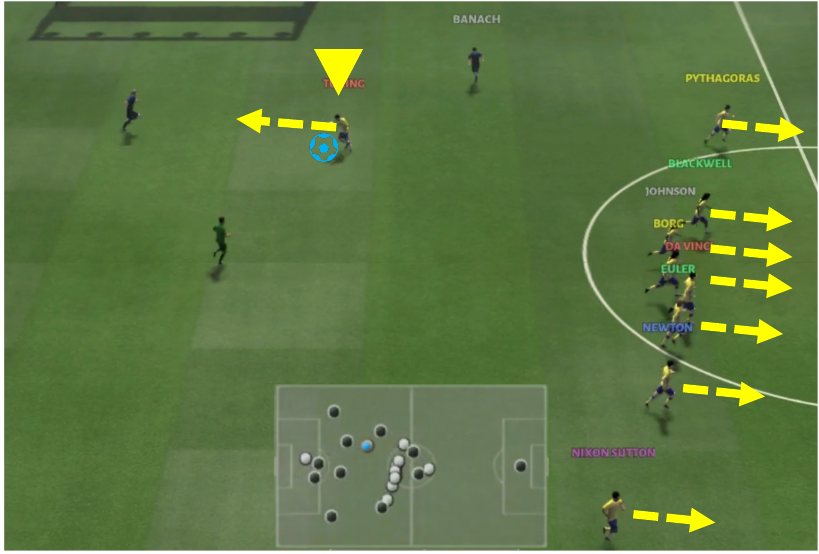}
    \label{figure0a}
 }
 \hspace{-0.3cm}
 \subfigure[]{
    \includegraphics[width=4.05cm]{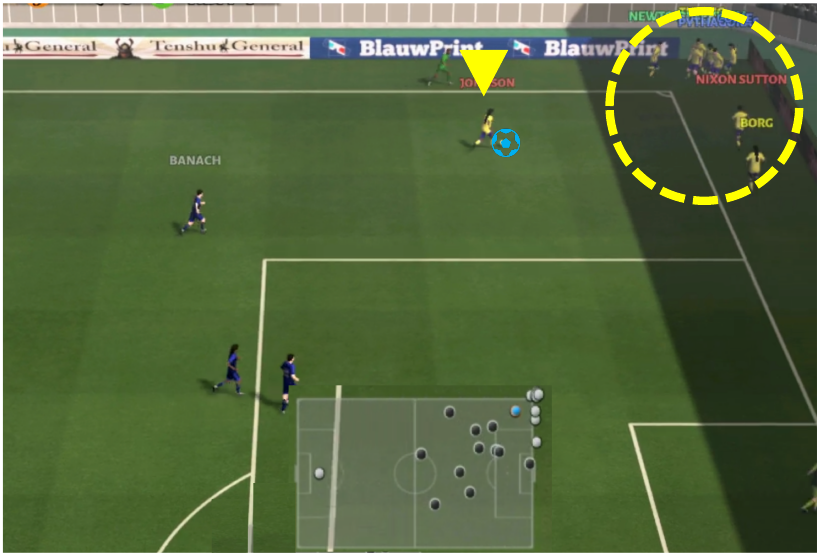}
    \label{figure0b}
 }
 \caption{Multi-agent policies trained by MAPPO and HAPPO in GRF 11 vs 11 scenario. (a) MAPPO: The non-ball-holding players of the yellow team do not form reasonable formation to occupy valuable space. (b) HAPPO: The non-ball-holding players of the yellow team lack meaningful positioning and movement to provide effective passing opportunities for the ball-holding player.}
 \label{figure0}
\end{figure}

Although state-of-the-art MARL algorithms can achieve high rewards in complex tasks with sparse rewards, their policies often diverges from human understanding. This discrepancy is reflected in the inconsistency between the learned policies in simulation environments and human common sense of the tasks. For instance, in Google Research Football (GRF) environment \cite{kurach2020google}, researchers typically hope to learn impressive tactical combinations which can inspire tactical choices in the real world. However, existing algorithms tend to hack the environment by finding and exploiting unintended shortcuts or loopholes to achieve higher rewards. \cite{bolander2018better, liu2023lazy}, which limits the effectiveness in real-world applications of MARL. As shown in Figure~\ref{figure0}, two typical unreasonable policies trained by MAPPO \cite{yu2022surprising} and HAPPO \cite{kuba2022trust} illustr the lack of structured formations and coherent passing coordination.

Therefore, it is a challenging issue that MARL algorithms is hard to learn policies which conform to human common sense. This issue mainly have three reasons: {\bf{(\romannumeral1)}} {\bf{Limitations of simulation environments.}} The complexity and diversity of the real world make it challenging to capture all relevant variables and dynamic changes. Thus, the inability to construct an environment model that is identical to the real world is an unavoidable problem. {\bf{(\romannumeral2)}} {\bf{Difficulty in modeling human common sense.}} Human common sense is often formed based on their rich experiences and comprehensive cognition, making it difficult to express accurately through simple mathematical formulas or rules. Consequently, we find it challenging to quantify and model human common sense using manually designed conventional reward functions, especially in complex and long-horizon multi-agent tasks. {\bf{(\romannumeral3)}} {\bf{Complexity and dynamic change of goals.}} In MARL, the typical objective for agents is to optimize specific and quantifiable performance metrics. However, human decision-making processes in complex tasks often involve dynamically changing goals across multiple stages. The simplified goal settings in traditional MARL fail to capture the complexity and diversity of human decision-making, resulting in a significant gap between the learned policies and human common sense. Some researchers leverage imitation learning \cite{hussein2017imitation} or reward shaping \cite{csikszentmihalyi2015intrinsic} to guide policies to align as closely as possible with human cognition. However, these methods have the drawback of relying on high-quality human behavior data and extensive domain knowledge.

Recent researches in foundation models have demonstrated notable achievements across various applications \cite{xiao2023chain,qian2023communicative,wang2023voyager,abi2024scaling}. These
models are useful because their pre-training encompasses
a vast amount of high-quality human behavior data and advanced understanding of various tasks. Therefore, the inherent capabilities of foundation models hold great potential and value in addressing the difficulty in modeling human common sense in complex MARL. Many researchers utilize the rich knowledge in visual-language models (VLM) to generate dense rewards for vision-based reinforcement learning tasks with sparse rewards \cite{adeniji2023language,rocamonde2023vision,baumli2023vision}. However, they primarily focus on single agent performing individual tasks, which does not address complex multi-agent tasks. In addition, to address the issues of complex and dynamically changing goals, researchers are dedicated to designing dense rewards that can capture the complexity of human decision-making, such as intrinsic rewards and potential-based rewards. However, existing designs still lack sufficient expressiveness and fail to model advanced semantics of common sense. Therefore, how to design comprehensive dense rewards that align with the common sense in human decision-making processes remains a significant challenge.

Based on the analyses, the paper investigates {\bf{V}}ision-based {\bf{GE}}neric {\bf{P}}otential {\bf{F}}untions {\bf{(V-GEPF)}} to facilitate hierarchical policy alignment in MARL. Policy alignment refers to the process where the learned policies not only pursue performance optimization, but also align with human common sense and cognitive patterns, whose significance lies in enhancing practicality and meaningfulness of the policies.

Specifically, to address the challenge of modelling human sense as a reward in complex MARL, we design a VLM-based generic potential function at the bottom layer. The generic potential function utilizes rich pretrained human knowledge within the VLM to guide policy learning towards alignment with human common sense. Then, recognizing that human decision-making involves dynamically changing goals, our method requires policies to flexibly adapt to uncertainty and change. Meanwhile, temporal information is crucial for understanding and aligning with the complex cognitive processes of human decision-making.
Therefore, based on the characteristics that visual large language models (vLLM) have broader human knowledge and the ability to handle more complex input information than VLM, we design a vLLM-based adaptive skill selection module. The module use more informative instructions, replayed videos, and reflection on the records of last potential function to adaptively select appropriate next potential function from a pre-designed pool, aiming to achieve a more comprehensive alignment with human common sense at the top layer. Furthermore, our method can be theoretically proven to not alter the optimal policy. The primary contributions of this work are as follows:

\begin{itemize}
    \item We propose a hierarchical vision-based reward shaping method for MARL, focusing on learning policies that better conform to human common sense, rather than simply optimizing for efficiency.
    \item To overcome the difficulty in modeling human common sense as rewards, we design VLM-based generic potential functions which utilize rich human knowledge from pre-trained data in VLM to guide policy alignment.
    \item Considering the dynamic objectives from multiple facets of the human decision-making process, we design a vLLM-based adaptive skill selection module to achieve a higher level and more comprehensive consistency between policies and human common sense.
\end{itemize}

\section{Related Works}
\subsubsection{Viusal-language models.}Foundational models trained on vast amounts of data can acquire broad and transferable representations of various data types, such as images and languages, making them applicable to a variety of downstream tasks \cite{wei2022emergent}. As type of foundation models, visual-language models (VLM) integrate linguistic and visual signals, playing a crucial role in fields that require processing of both modalities \cite{wang2021simvlm,ju2022prompting}. There are diverse applications of VLM in reinforcement learning (RL). For instance, VLM can be used as reward functions \cite{dang2023clip}, promptable representation learners \cite{chen2024vision}, and for data augmentation based on hindsight relabeling \cite{sumers2023distilling}. The applications demonstrate wide applicability and robust capabilities of VLM, providing effective tools and new perspectives for addressing complex RL challenges.

\subsubsection{VLM as RL rewards.}Currently, a promising application direction is using VLM to generate dense rewards for RL tasks, especially in scenarios with sparse rewards \cite{fu2024furl,rocamonde2023vision,wang2024rl}. Cui et al. \cite{cui2022can} utilize the pretrained CLIP to provide image-based rewards for robotic manipulation tasks. Mahmoudieh et al. \cite{mahmoudieh2022zero} successfully apply the fine-tuned CLIP as a language-described reward model for robotic tasks. Sontakke et al. \cite{sontakke2024roboclip} use VLM in a robotic environment to provide reward signals for RL agents, primarily defining rewards through video demonstrations. However, existing methods primarily focus on solving specific single-agent tasks, where the design of the reward function tends to be relatively simple and direct. Moreover, these methods are mostly implemented in vision-based environments. In contrast, multi-agent systems involve more complex interactions and collaboration mechanisms, making dense reward guidance more challenging.

\subsubsection{Potential-based rewards.}In MARL tasks with sparse rewards, the introduction of additional dense rewards to represent domain-specific expert knowledge has become a common practice to accelerate training \cite{mguni2021ligs,ma2022elign}. However, it soon became apparent that if used improperly, the dense rewards might alter the optimal response, which directly affects the performance of the policies \cite{csimcsek2006intrinsic}.Ng et al. \cite{ng1999policy} demonstrate that the optimal policy do not change if the dense reward is designed as $F(s,s')=\gamma\phi(s')-\phi(s)$. Devlin et al. \cite{devlin2012dynamic} further prove that time-varying potential-based rewards $F(s,t,s',t')=\gamma\phi(s',t')-\phi(s,t)$ do not alter the Nash equilibrium in multi-agent problems. Grzes \cite{grzes2017reward} extend the proof to episodic tasks that terminate after a certain final time step $N$. Chen et al. \cite{chen2022knowledge} design potential-based rewards in GRF to be positively correlated with the distance between the ball and the goal. Zhang et al. \cite{zhang2023multiexperience} design the rewards that are proportionally linked to the remaining health of a unit in StarCraft II \cite{samvelyan2019starcraft}. However, existing methods are often based on simple formulas or rules, making it difficult to comprehensively represent human common sense.

\section{Background}
\subsubsection{Partially observable Markov decision process.}A multi-agent reinforcement learning problem in vision-based tasks can be formally defined as a partially observable Markov decision process (POMDP) \cite{puterman1990markov}. A POMDP is represented as a tuple: $<\mathcal{N},\mathcal{S},\mathcal{A},\mathcal{P},r,\mathcal{G},\mathcal{O},\gamma>$, where $|\mathcal{N}|=N$ is the number of agents, $s\in\mathcal{S}$ is the state space, $\gamma\in[0,1]$ is a discount factor. At each time step $t$, the agent $i\in \mathcal{N}$ chooses its action $a_t^i\in\mathcal{A}$ based on its observations $o_t^i\in\mathcal{O}$. The actions of all agents constitute the joint action space $a\in\mathcal{A}^N$. The state and joint action make up a joint reward $r(s,a,s')$ based on the state transition function $\mathcal{P}(s'|s,a)$. The team's objective is to maximize the expected discounted return $G=\sum_{t=0}^{\infty}\gamma^tr(s_t,a_t,s_{t+1})$.

\subsubsection{CLIP models.}VLM \cite{zhang2024vision} is a typical foundation model that can handle sequences incorporating both language inputs $l \in \mathcal{L}^{\leq n}$ and visual inputs $v \in \mathcal{V}^{\leq m}$. $\mathcal L$ represents a finite alphabet encompassing strings of length no greater than $n$, while $\mathcal V$ represents the space for 2D RGB images whose length do not exceed $m$. Contrastive language-image pretraining model (CLIP) \cite{radford2021learning} is a representative VLM which trains by aligning image and text embeddings in the latent space. Specifically, the CLIP model consists of a text encoder $\tau_L$ and an image encoder $\tau_I$, both of which are mapped to a same latent space $\mathbb{R}^k$. Typically, the encoders are trained by minimizing the cosine distance between embeddings for pairs of images and languages.

\begin{figure*}[t]
\centering
\includegraphics[width=0.88\textwidth]{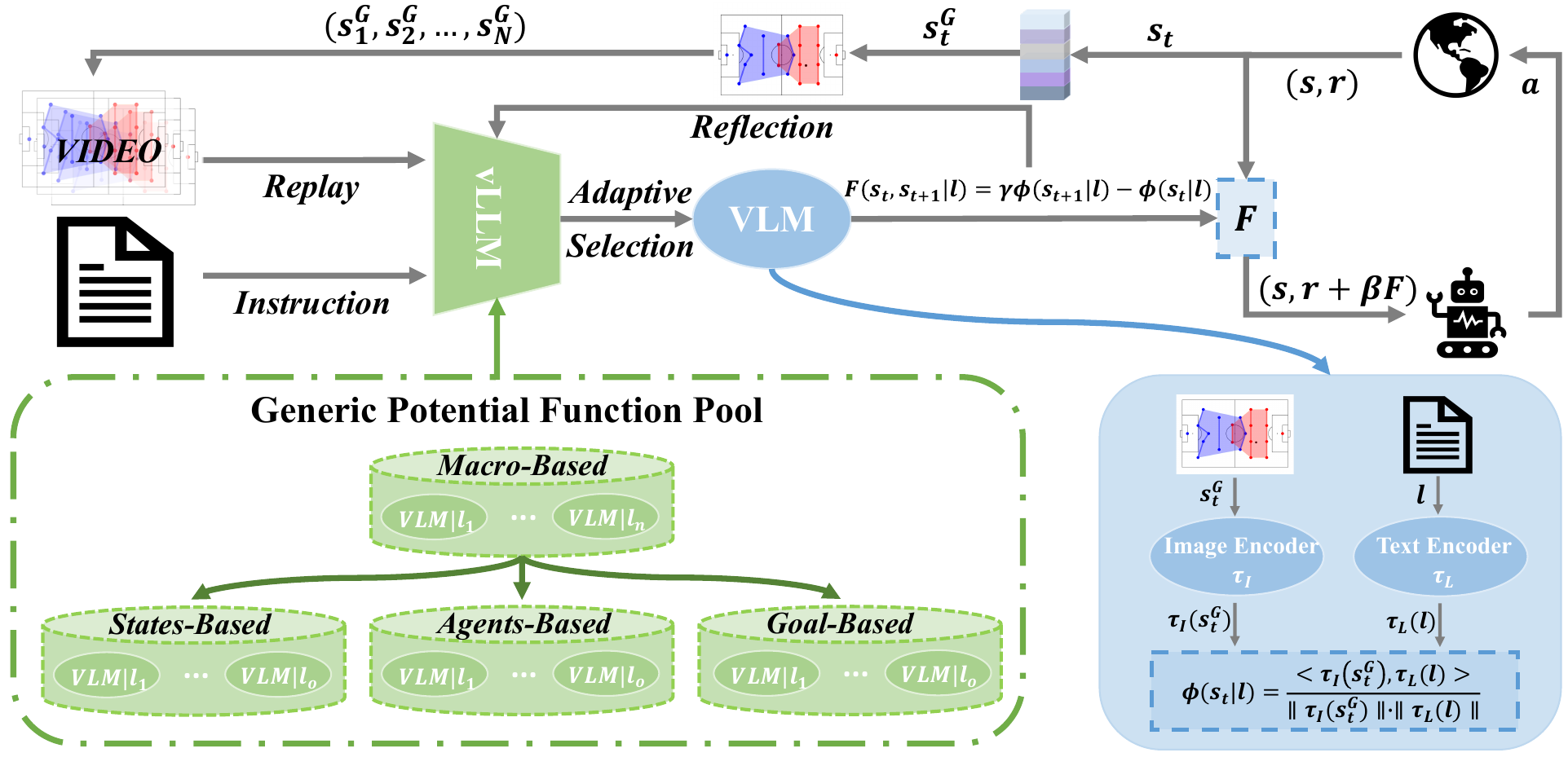} 
\caption{Framework of the V-GEPF. The state's image representation $s_t^G$ and the human instructions $l$ are input separately into image encoder and text encoder. The cosine distance computed from the outputs is designed as a generic potential function $\phi(s_t|l)$, which is weighted and combined with the original environmental rewards to guide the learning of policies. Furthermore, the video replay of last episode, the initial human instructions, information about potential function pool, and the reflection on the records of last potential function are fed into vLLM to adaptively select appropriate generic potential function at various training phase.}
\label{fig2}
\end{figure*}
\section{Methodology}
The goal of this paper is to explore how to learn policies that are more aligned with human common sense in complex MARL with sparse rewards. We propose a hierarchical vision-based reward shaping method named V-GEPF, which utilizes rich human knowledge from foundation models to guide policy alignment from different levels. The algorithm framework is described in Appendix~\ref{app: algorithm}.

\subsection{VLM as Generic Potential Functions}
Guiding MARL policies to align with human common sense is a challenging problem. Human common sense is typically formed based on rich experiences and comprehensive cognition of specific tasks, making it challenging to be modelled and quantified through simple formulas or rules, especially in complex and long-horizon multi-agent tasks.

Recent researches have shown effectiveness of additional dense rewards, such as potential-based rewards, to guide policy alignment \cite{chen2022knowledge,zhang2023multiexperience}. However, although these dense rewards are useful in certain tasks, they do have some limitations: {\bf{(\romannumeral1) Insufficient expressiveness.}} Manually designed dense rewards are overly simplistic, failing to fully capture the multifaceted goals and high-level semantic information of human common sense in complex tasks. {\bf{(\romannumeral2) Limited understanding ability.}} Dense rewards often rely on the intuition and experience of experts, which may overlook complex behavioral dynamics and environmental factors in tasks, leading to a lack of in-depth and comprehensive understanding of complex tasks. {\bf{(\romannumeral3) Lack of adaptability and flexibility.}} Dense rewards are typically fixed, which are struggle to cope with dynamic changes in the environment. These limitations make it difficult to effectively align multi-agent policies with human common sense in complex decision-making tasks through existing dense rewards.

The VLM possesses inherent exceptional capabilities due to the rich expert knowledge it has acquired from training datasets encompassing various complex tasks. Therefore, we design VLM as generic potential functions to guide policies in exploring directions that align with human common sense.

\subsubsection{Instruction-conditioned potential function.}
To design VLM-based generic potential functions, we first introduce an instruction-conditioned potential function. By incorporating the concept of instructions into the potential function, we lay the groundwork for exploring the alignment of VLM-guided policies with human common sense.

\begin{myDef}
    \label{1}
    An \textbf{instruction-conditioned potential function} $\phi(s|l):S\times L\rightarrow R$ is a mapping from the current state $s$ to its latent value $\phi$, given the language instruction $l$.
\end{myDef}
The instruction-conditioned potential function does not directly depend on the agents' actions, but is determined by the state under the given instructions. The instructions reflect the guidance of human common sense. It is worth noting that when considering trajectories with a finite time step, the final step $N$ has the terminal state, whose potential value should be set to zero \cite{wierstra2008episodic}:
\begin{equation}
\label{potential1}
\phi(s_t|l) = \begin{cases} 
0 & \text{if } t = N \\
\phi(s_t|l) & \text{otherwise.}
\end{cases}
\end{equation}

\subsubsection{VLM as a generic potential function.}
The pre-training data of VLM includes a vast repository of high-quality human understanding and cognition related to complex tasks. As a result, VLM has significant potential and value in overcoming the challenges of modeling human common sense in complex MARL tasks.

Contrastive language-image pretraining model (CLIP) \cite{radford2021learning} is a typical efficient VLM which is designed as generic potential functions in this paper. CLIP consists of an image encoder, $\tau_I$, and a text encoder, $\tau_L$. These encoders map images and text to embeddings in the same latent space. They are trained on a dataset containing a large number of image-text pairs by minimizing the cosine distance between the corresponding embeddings. Therefore, images and texts with similar semantics can be mapped to similar embeddings. We first convert the environmental state $s_t$ into an image $s_t^G$, enabling the CLIP to more intuitively understand the current situational context. Subsequently, the image and the corresponding instruction are fed into the image encoder and text encoder respectively, producing two embeddings $\tau_I(s_t^G)$ and $\tau_L(l)$. These embeddings reflects the semantics of instruction and current state. To align the policy with human common sense, the VLM-based generic potential function is designed as the cosine distance between the state and instruction embeddings:
\begin{equation}
    \label{cosine}
    \phi(s_t | l) = \frac{\langle \tau_I(s_t^G), \tau_L(l) \rangle}{\|\tau_I(s_t^G)\| \cdot \|\tau_L(l)\|}.
\end{equation}
\noindent



Then, based on Def. \ref{1}, we subsequently design {\bf{generic potential-based reward}} as:
\begin{equation}
    \label{poten_f}
    F(s_t,s_{t+1}|l)=\gamma\phi(s_{t+1}|l)-\phi(s_t|l),
\end{equation}
\noindent
where $s_t$ and $s_{t+1}$ represent the current and next state respectively, $\gamma$ is the discount factor of MARL to ensure the designed rewards do not alter the optimal policy (detailed proofs can be found in Appendix~\ref{app:proof}). With the cross-modal representation capabilities of CLIP, this generic potential-based reward can enhance the agents' understanding of the semantic of state. 
Instruction $l$ is set to represent a ideal state that align with human common sense, e.g. `the home players show cooperative attacking tactics'. Detailed instructions $l$ are provided in Appendix~\ref{app: prompt}. The visualization details of states $s_t^G$ are presented in Appendix~\ref{app: CLIPimage}.

Since there is an original reward $r_{env}$ provided by the environment, the total reward is described as:
\begin{equation}
    \label{poten_r}
    R(s_t,s_{t+1}) = r_{env}(s_t,s_{t+1}) + \rho \cdot F(s_t,s_{t+1}|l),
\end{equation}
\noindent
where $\rho$ represents a scalar coefficient to balance the generic potential-based reward with the environmental reward.

It can be theoretically demonstrated that incorporating the generic potential-based reward ensures that the Nash equilibrium in multi-agent problems remains unchanged. Detailed proofs are presented in Appendix~\ref{app:proof}.

However, the decision-making process in complex long-horizon tasks often involves multiple dynamically changing goals, which means that the guiding direction of the policy needs to adaptively adjust at different stages. This is a key characteristic of human decision-making process. In this context, generic potential-based reward with a fixed instruction is inadequate, as it cannot comprehensively and accurately quantify the diversity and dynamics of these complex goals. Meanwhile, the lack of temporal information limits the understanding and aligning with the complex cognitive processes involved in human decision-making. These limitations hinder our policies' ability to effectively align with human common sense to some extent. To enhance flexibility and adaptability in our method, we design a vLLM-based adaptive skill selection module.

\subsection{vLLM-based Adaptive Skill Selection}
In complex long-horizon tasks, human decision-making process can adapt to dynamic changes and involve multidimensional objectives. Taking football match as an example, a team's policies will flexibly adjust according to the progress of the match and the current situation. At the beginning of the match or when the advantage is not significant, a team often employ more aggressive offensive tactics, focusing on coordination among players to achieve rapid passing or multi-player attacks. When holding a significant lead, the team pay more attention to maintaining the stability of their formation and controlling ball possession to preserve their advantages \cite{harrop2014performance}. 


To align the decision-making process of MARL with human cognition, the designed rewards must be capable of adaptively and flexibly adjusting to the uncertain and changing circumstances. Leveraging the advanced understanding and representation capabilities of the visual large language model (vLLM), we propose a vLLM-based adaptive skill selection module. This module dynamically selects different skills (VLMs with pre-defined instructions) at various stages of training, allowing MARL agents to adjust their behaviors based on the evolving game context.

We design a generic potential function pool emulating the thought processes involved in human decision-making. As researched by Simon \cite{simon1955behavioral}, in the decision-making process, humans first attempt to understand the macro environment, then turn their attention to the specific details at the micro-level, which includes the analysis of individual behaviors and the pursuit of goals. Thus, different generic potential functions are designed from macro and micro perspectives to guide policies exploring towards the respective desired directions. At the macro level, the potential functions focus on the overall situation (such as team advantage) in order to set macro objectives in complex environments. At the micro level, we analyze from three perspectives: states-based, agents-based, and goal-based.
\begin{itemize}
    \item States-based skills focus on analyzing microscopic state information, such as the local advantage around a key agent.
    \item Agents-based skills focus on the behaviors and interactions of the agents to uncover the complex dynamics of their actions and relationships, such as the level of coordination between the agents.
    \item Goal-based skills aim to maintain a keen focus on specific objectives, such as the status of key targets.
\end{itemize}

These different types of skills enable the policies to make more refined and targeted choices and adjustments in various contexts, promoting a more comprehensive alignment with human common sense. We provide a more detailed description of the designed instructions in Appendix \ref{app: prompt}.

Furthermore, information on the temporal dimension is crucial for capturing the continuity and evolution of policy learning. Thus, we present the entire episode's states in a video replay to provide temporal information as input to the vLLM.

Overall, the vLLM-based adaptive skill selection module and the VLM-based generic potential function hierarchically construct a vision-based generic potential function. The vLLM receives the video replay of the last episode, initial human instructions, information about the potential function pool, and the reflection on records of the previous potential function. Using these information, a VLM-based potential function is selected from the pool, providing multi-angle, multi-level guidance for policy alignment.
The dialogue examples of this module are provided in Appendix~\ref{app: dialogue}.
\section{Experiments}




\subsection{Experimental Setup}

All experiments are implemented in a typical long-horizon multi-agent benchmark Google Research Football (GRF) \cite{kurach2020google}, which simulates real football matches and serves as a widely used benchmark for researchers in football game AI. We evaluate our method on GRF full-field (11 vs 11) tasks with different difficulty levels. In the full-field tasks, all 11 players of home team's are controlled by RL policies, while the opposing 11 players are managed by a built-in script with adjustable difficulty settings. Each player has a 19-dimensional discrete action space, with all players sharing global state information. This constitutes a vast exploration space. We set the reward to be given only when a goal is scored, and each episode can be up to 3000 steps long, which constitutes a sparse reward problem. The full-field tasks are sufficiently challenging, making it very difficult for MARL to learn good collaborative policies within them.

V-GEPF is implemented based on MAPPO and compared against MAPPO \cite{yu2022surprising}, IPPO \cite{de2020independent}, and HAPPO \cite{kuba2022trust}. During training, all algorithms set 32 workers to sample training data in parallel, with each worker sampling 3,000 steps per epoch. The training ends after 300 epochs, resulting in approximately 28 million steps. Additionally, V-GEPF adaptively selects skills every 50 epochs. For the VLM, we use CLIP-RN50\footnote{\url{https://github.com/openai/CLIP}}, for the vLLM, we employ MiniCPM-Llama3-V\footnote{\url{https://huggingface.co/openbmb/MiniCPM-Llama3-V-2_5}}. To further evaluate V-GEPF, we introduce a baseline called MAPPO-xT, which also utilizes potential-based rewards. Using data from the FIFA World Cup 2018\footnote{\url{https://figshare.com/collections/Soccer_match_event_dataset/4415000}}, we calculate expected threat (xT) values for different positions on the field. In MAPPO-xT, the mapping from positions to xT values (the higher the better) serves as a potential function to guide MAPPO's policy learning. Calculation details of xT can be found in Appendix~\ref{app: xT}. The hyperparameters of all algorithms can be found in Appendix~\ref{app: hyperparameters}.

\subsection{Careful Comparison to SOTA}
\begin{figure}[htbp]
 \centering
 \subfigure[11 vs 11 easy.]{
    \includegraphics[width=3.95cm]{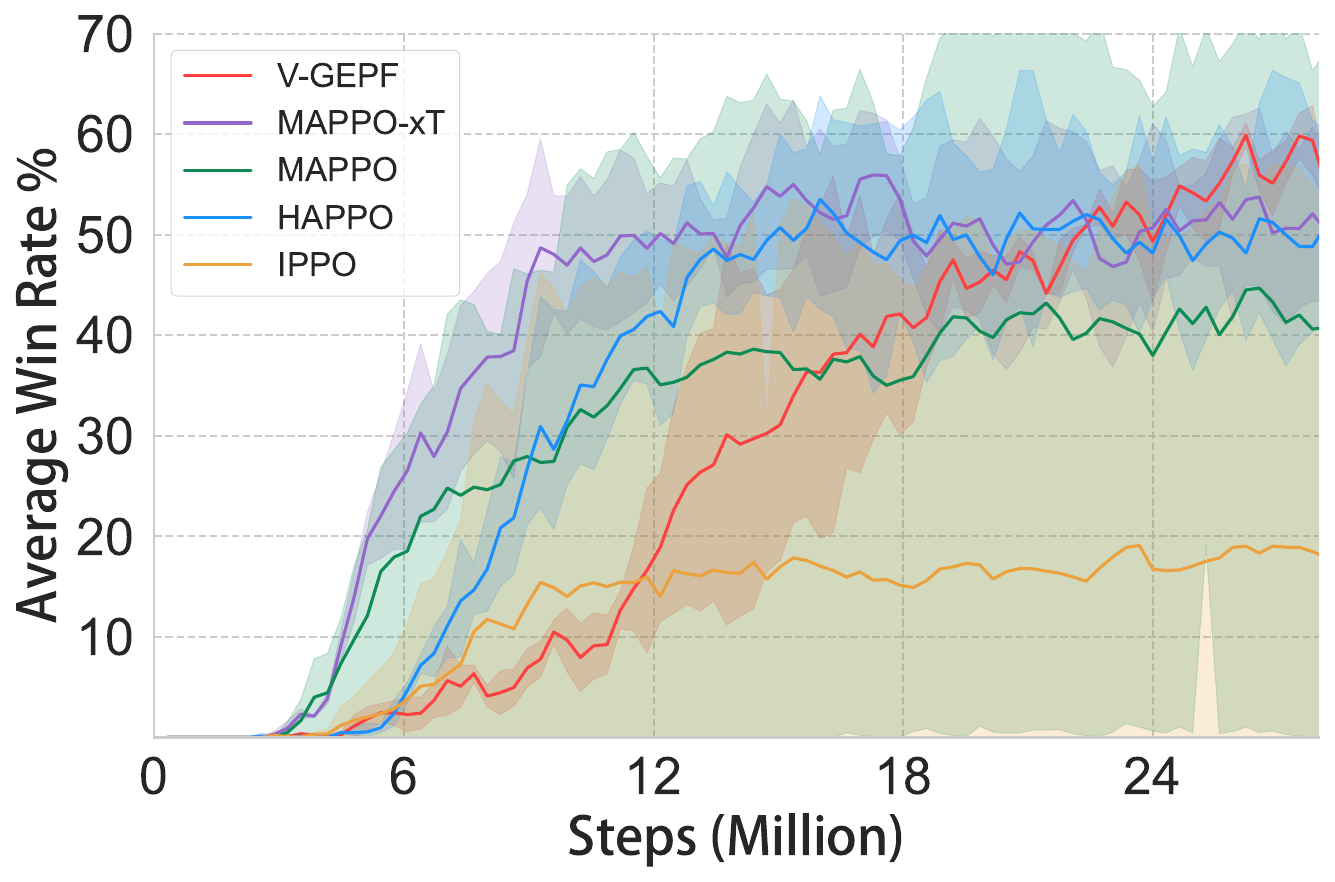}
    \label{figure_easy}
 }
 \subfigure[11 vs 11 hard.]{
    \includegraphics[width=3.95cm]{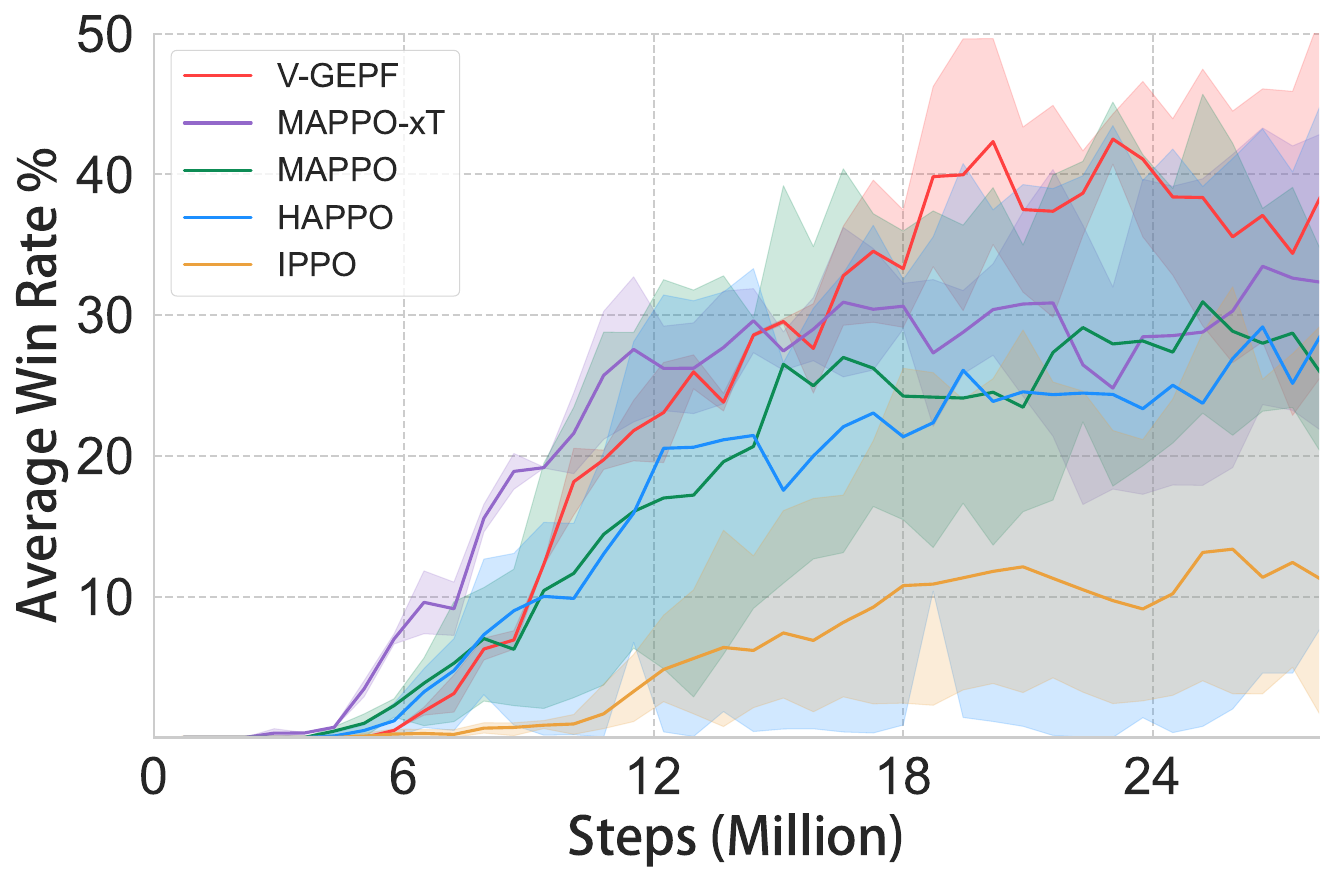}
    \label{figure_hard}
 }
 \caption{Average win rate curves during training in GRF 11 vs 11 scenario.}
 \label{figure3}
\end{figure}

We compare the average win rate curves of V-GEPF and state-of-the-art (SOTA) baselines during training. As illustrated in Figure~\ref{figure_easy}, in the 11 vs. 11 easy task, V-GEPF outperforms all baselines in terms of win rate, significantly surpassing IPPO. While HAPPO, MAPPO, and MAPPO-xT exhibits an advantage in convergence speed, their final win rates are lower than that of V-GEPF. The slower convergence speed is attributed to the potential-based reward guiding the policy to explore cooperative policy that align with human common sense, e.g. dribbling, keep formation, rather than solely focusing on scoring goals.

In the 11 vs. 11 hard task, as illustrated in Figure~\ref{figure_hard}, V-GEPF significantly outperforms all baselines in terms of win rate. Both MAPPO, MAPPO-xT, and HAPPO's average win rate curves plateau around 15 million steps. By visualizing SOTA policies, we observe that they have learned strange policies (as shown in Figure~\ref{figure0}): one player dribbles the ball toward the goal while the other players fail to cooperate, instead running toward the opponent's sideline or congregating at the corner or edge of the opponent's field. This represents a local optimum that RL can easily discover. However, once a policy is trapped in this local optimum, it becomes challenging to explore alternative policies due to the vast exploration space of this task. In contrast, V-GEPF benefits from the adaptive guidance provided by the vLLM and VLM, enabling the policy to avoid getting stuck in this local optimum. As a result, V-GEPF eventually achieves a higher win rate compared to the baselines.

\begin{figure}[htbp]
  \centering
    \subfigure[MAPPO]{              
        \includegraphics[width=3.8cm]{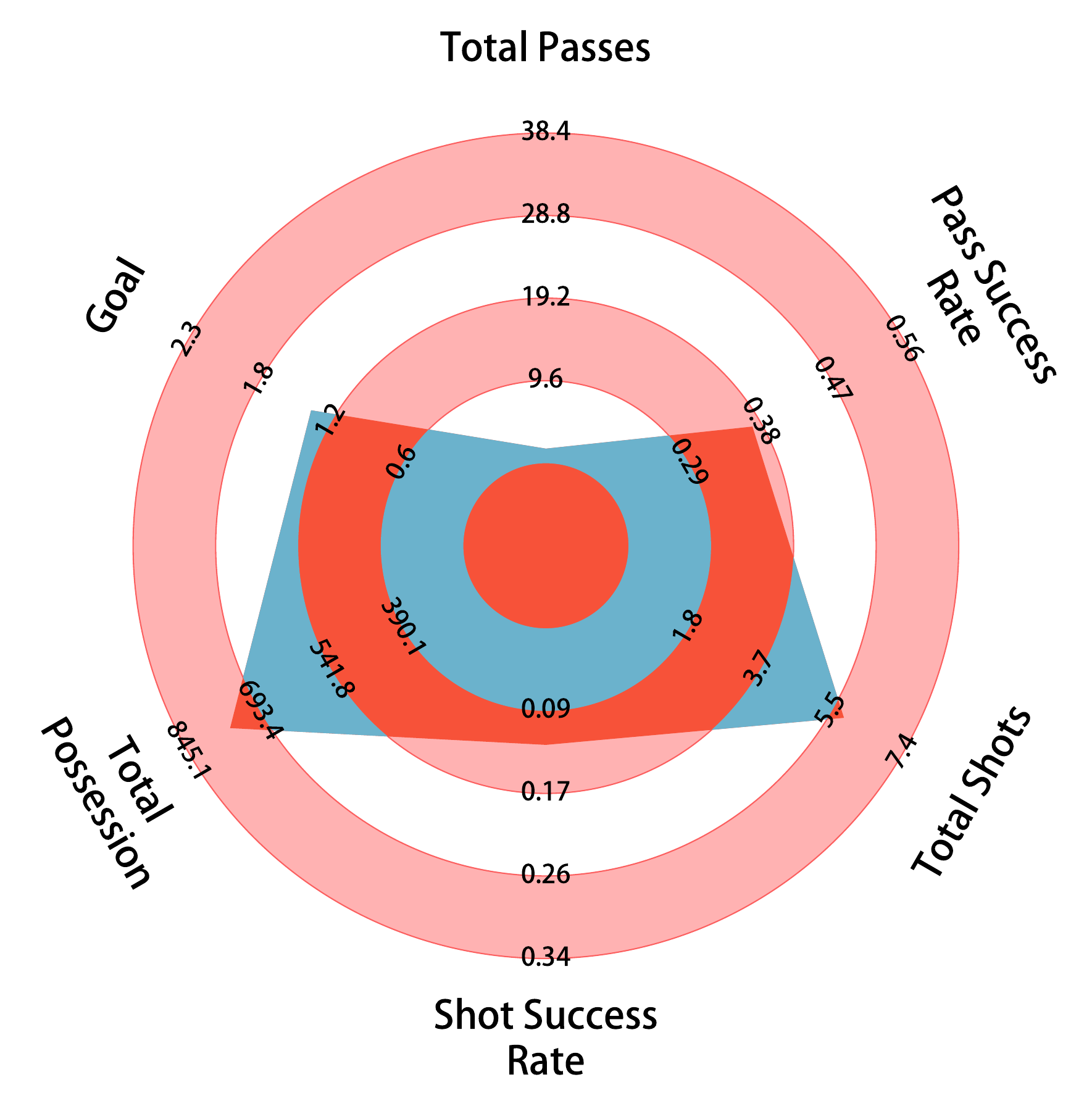}
        \label{fig:radar-mappo}}
    \subfigure[V-GEPF]{
        \includegraphics[width=3.8cm]{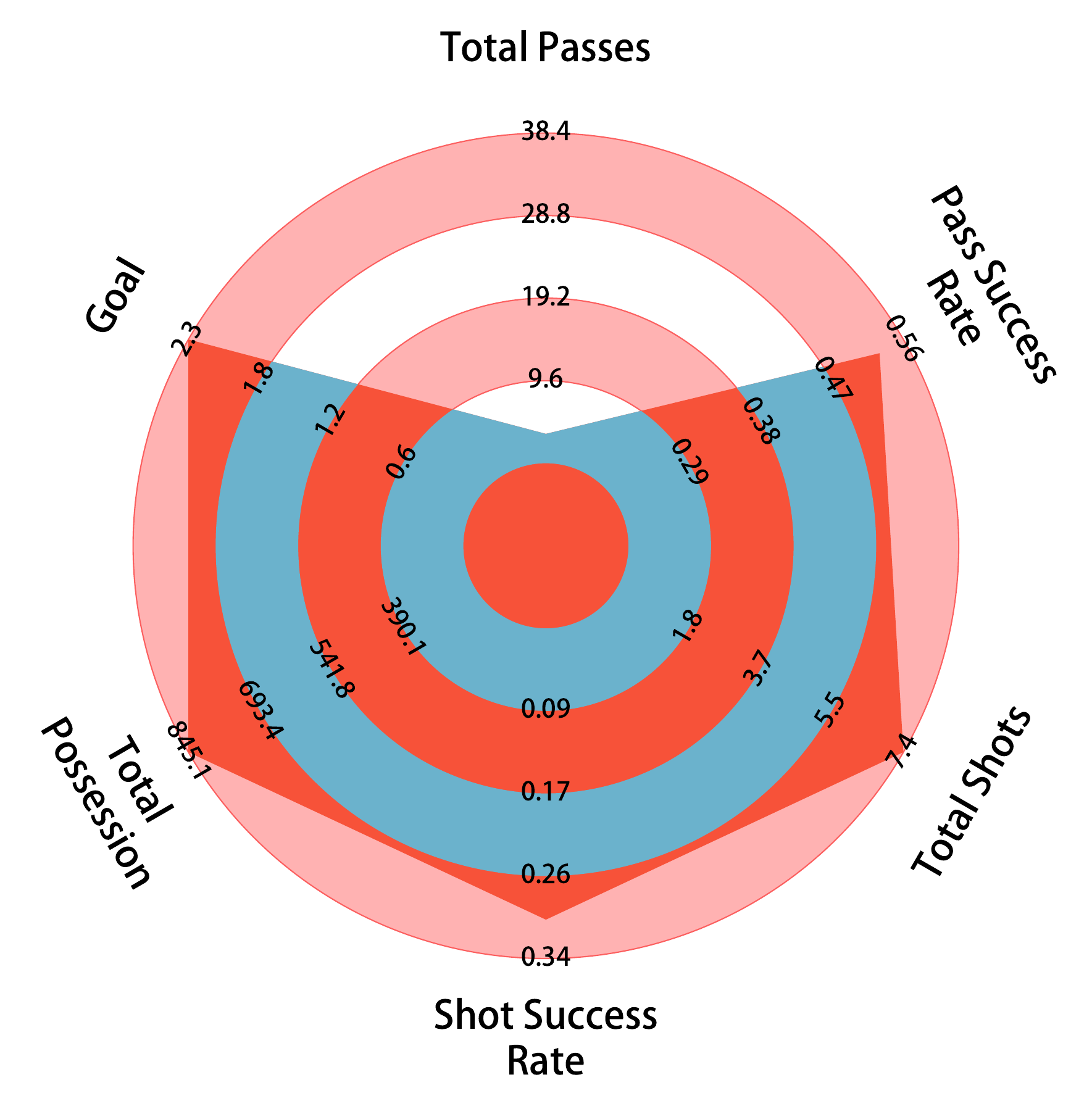}
        \label{fig:radar-vgepor}}
    \caption{Comparison of policy styles trained with MAPPO and MAPPO enhanced by V-GEPF.}
    \label{fig:radar-exp}
\end{figure}

We further use radar maps for a fine-grained comparison between the policies of MAPPO and V-GEPF. As shown in Figure~\ref{fig:radar-exp}, the six dimensions in the radar chart characterize how well a policy performs beyond the average win ratio, which can comprehensively reflect human common sense. V-GEPF is superior to MAPPO in all aspects, indicating that V-GEPF leads to a policy that is more aligned with human cognition.

\subsection{Potential Function Analysis}
\begin{figure}[h!tbp]
    \centering
    \includegraphics[width=0.7\linewidth]{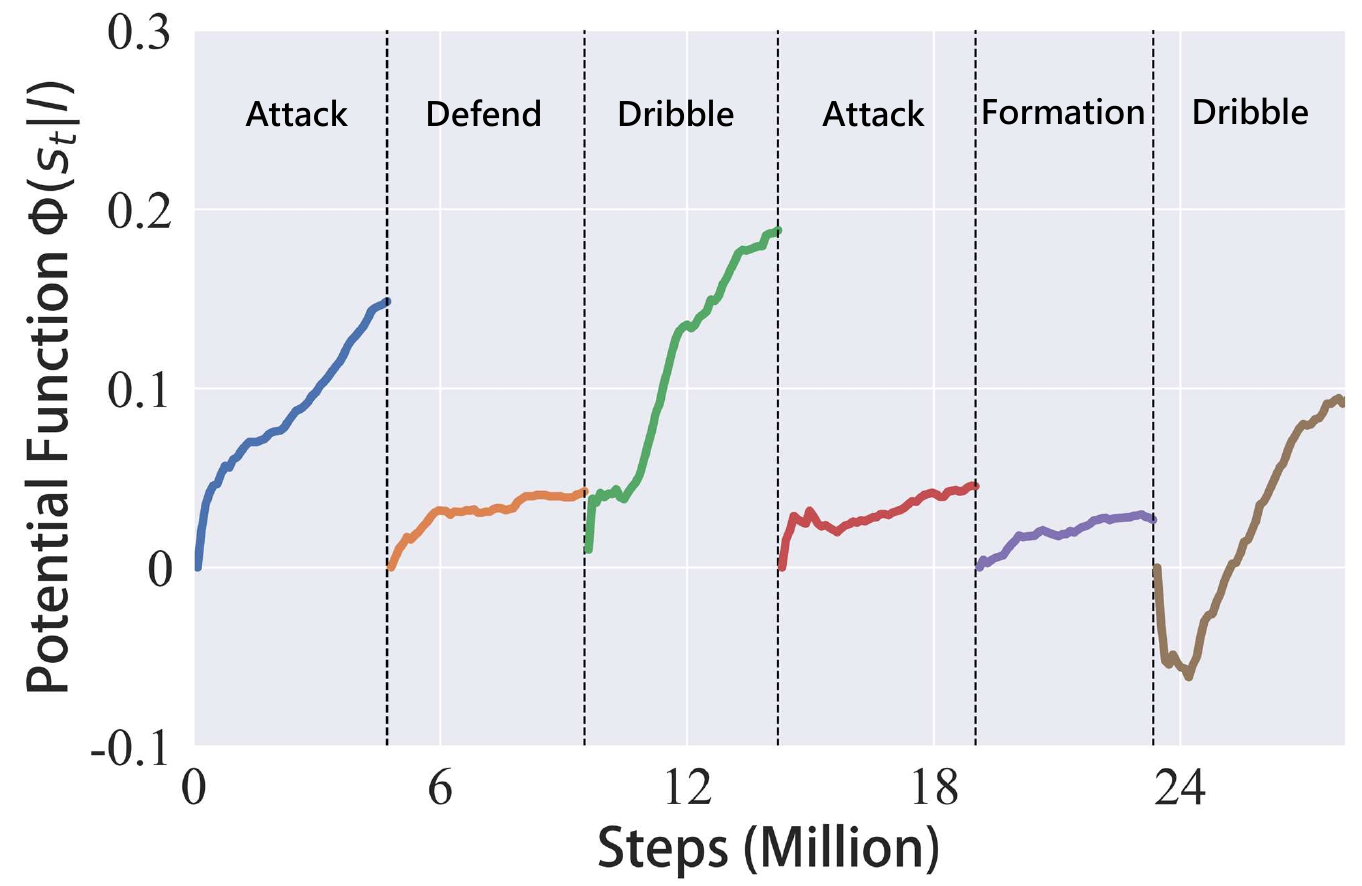}
    \caption{Potential function curves during training. Six VLM-based potential functions are selected sequentially by a vLLM according to replayed videos.}
    \label{fig:exp_potential}
\end{figure}

We select one run of V-GEPF, and record the average value of the potential function per epoch to analyze in detail how the potential function curve changes during training. After each skill selection, we use the first epoch to calculate the mean and standard deviation of the potential function and normalize it, so the potential function is always zero at the beginning. As shown in Figure~\ref{fig:exp_potential}, each potential function shows an upward trend, which indicates that VLM is able to guiding policy learning given different instruction. From the selection of skill, we find that vLLM can flexibly adapt skills to address the deficiencies of the current policy.

\begin{figure*}[t!]
\centering
\includegraphics[width=0.85\textwidth]{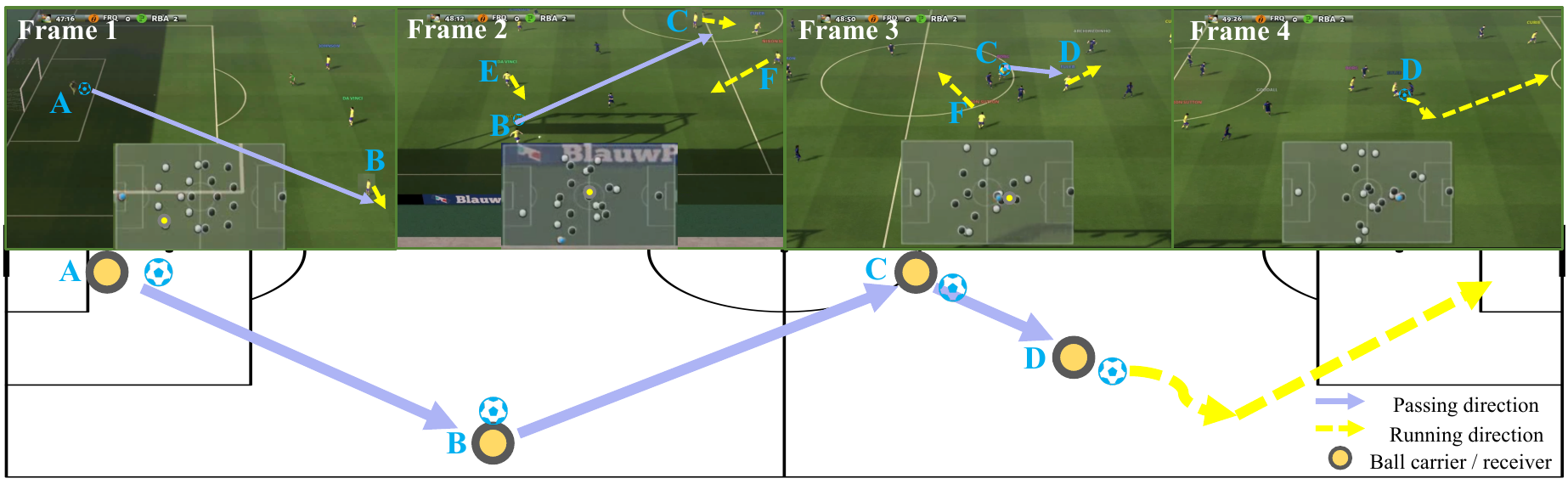} 
\caption{A visual policy analysis of the V-GEPF method in an offensive phase. The visualization shows that the agents have organized positioning and coherent passing combinations, which aligns with human common sense in real football matches.}
\label{visualization}
\end{figure*}

\begin{figure}[t!]
  \centering
    \subfigure[Attack]{              
        \includegraphics[width=3.8cm]{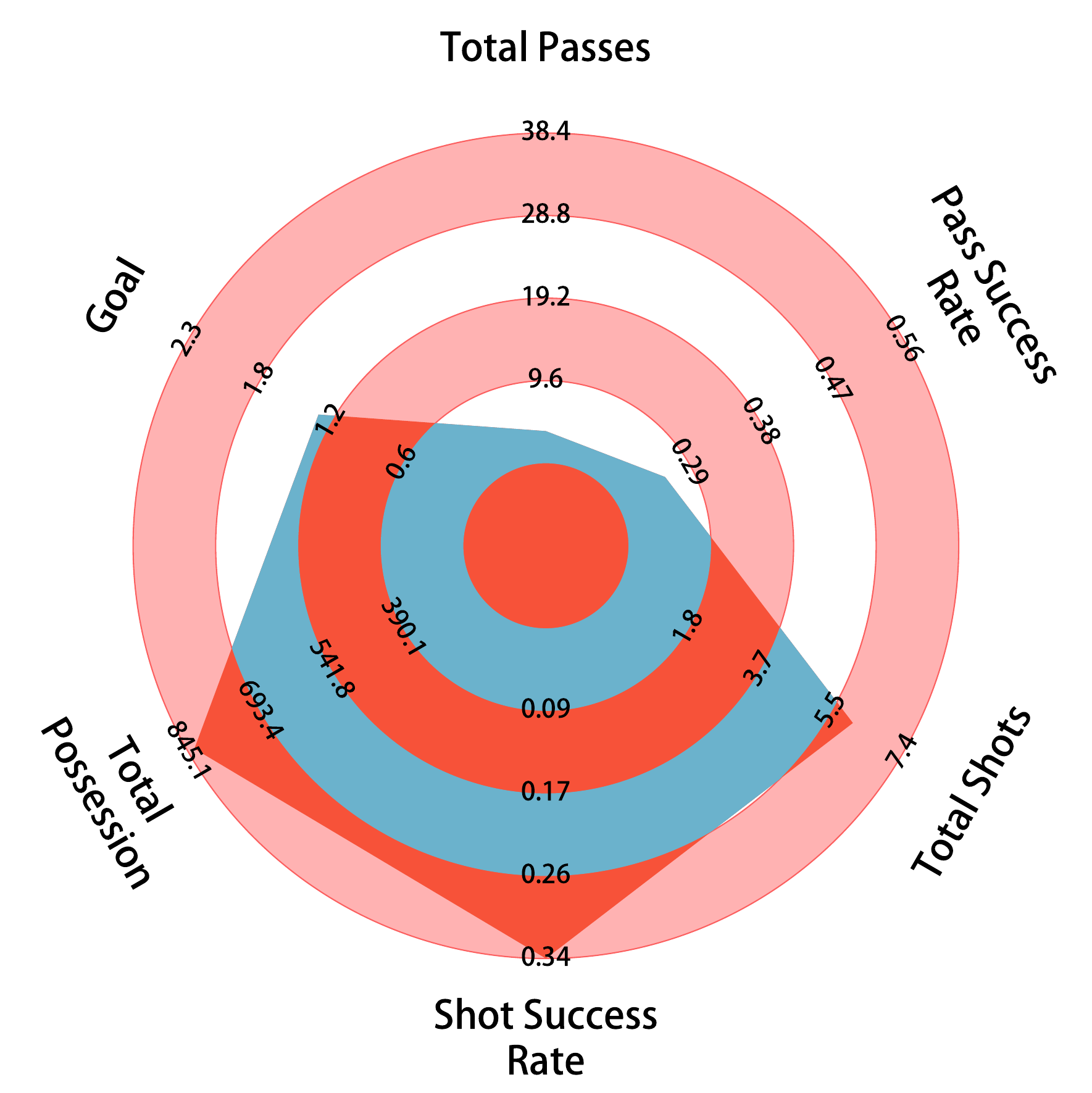}
        \label{fig:attack}}
    \subfigure[Defend]{
        \includegraphics[width=3.8cm]{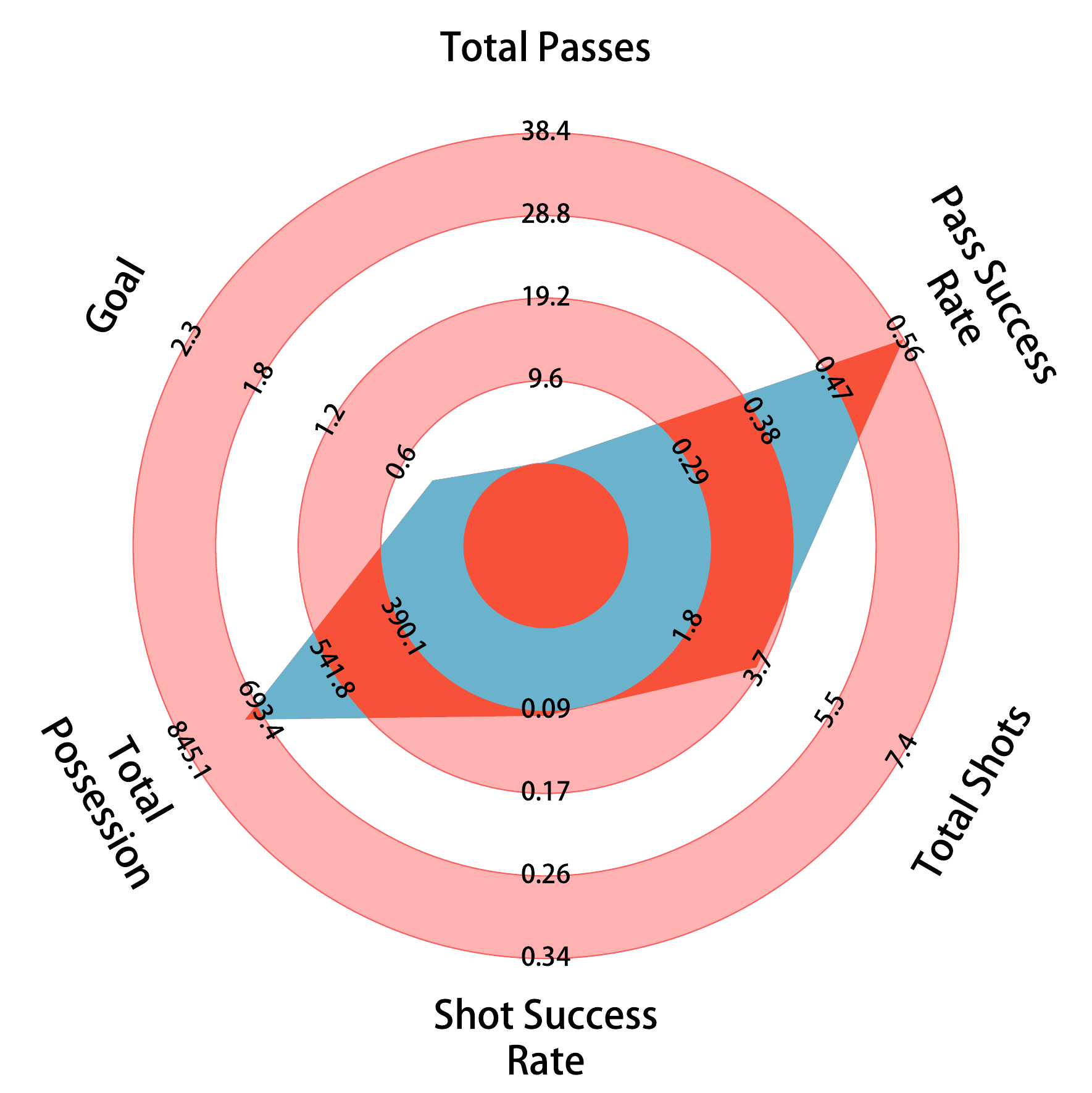}
        \label{c}}
    \subfigure[Dribble]{
        \includegraphics[width=3.8cm]{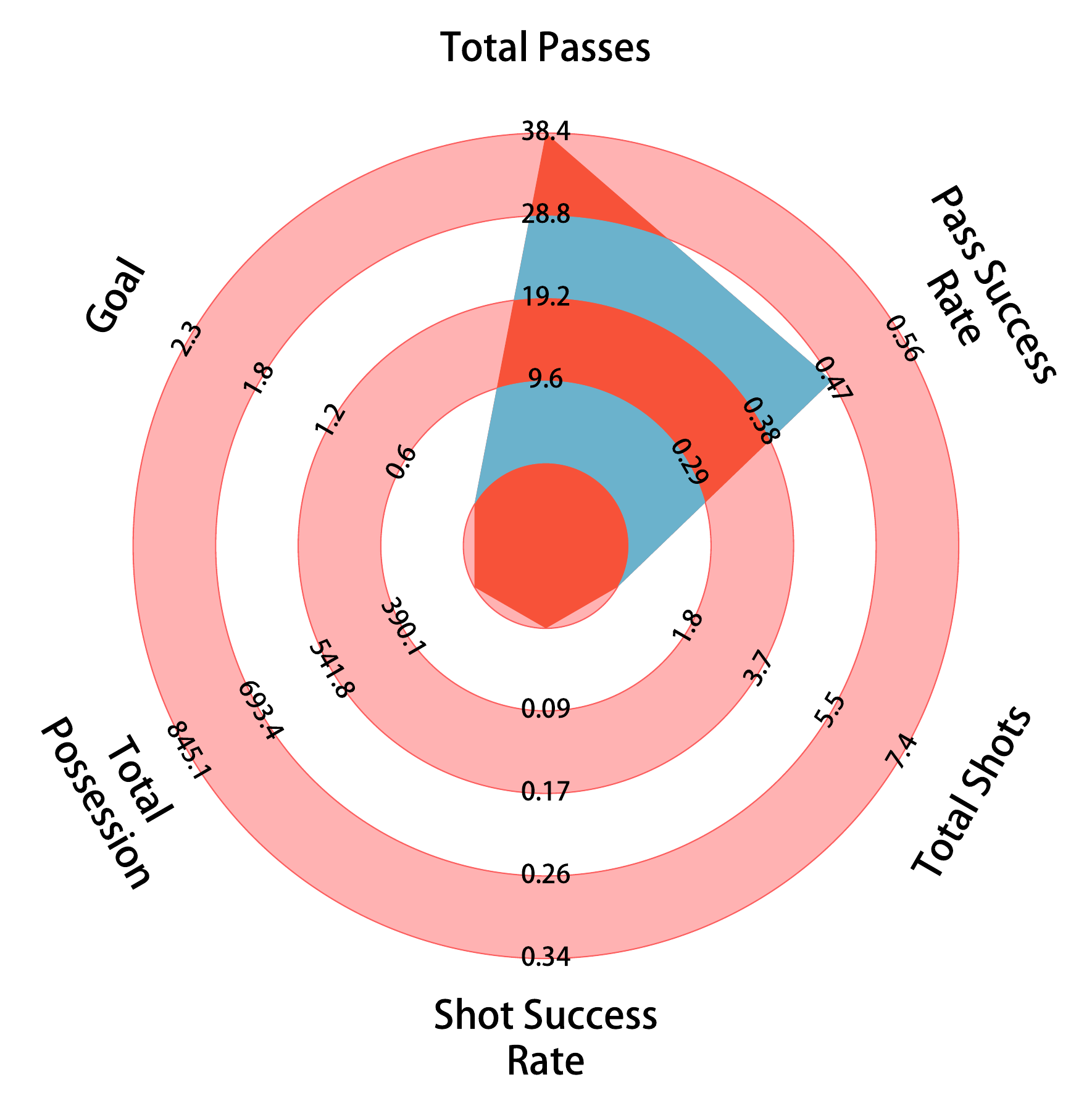}
        \label{fig:dribble}}
    \subfigure[Formation]{
        \includegraphics[width=3.8cm]{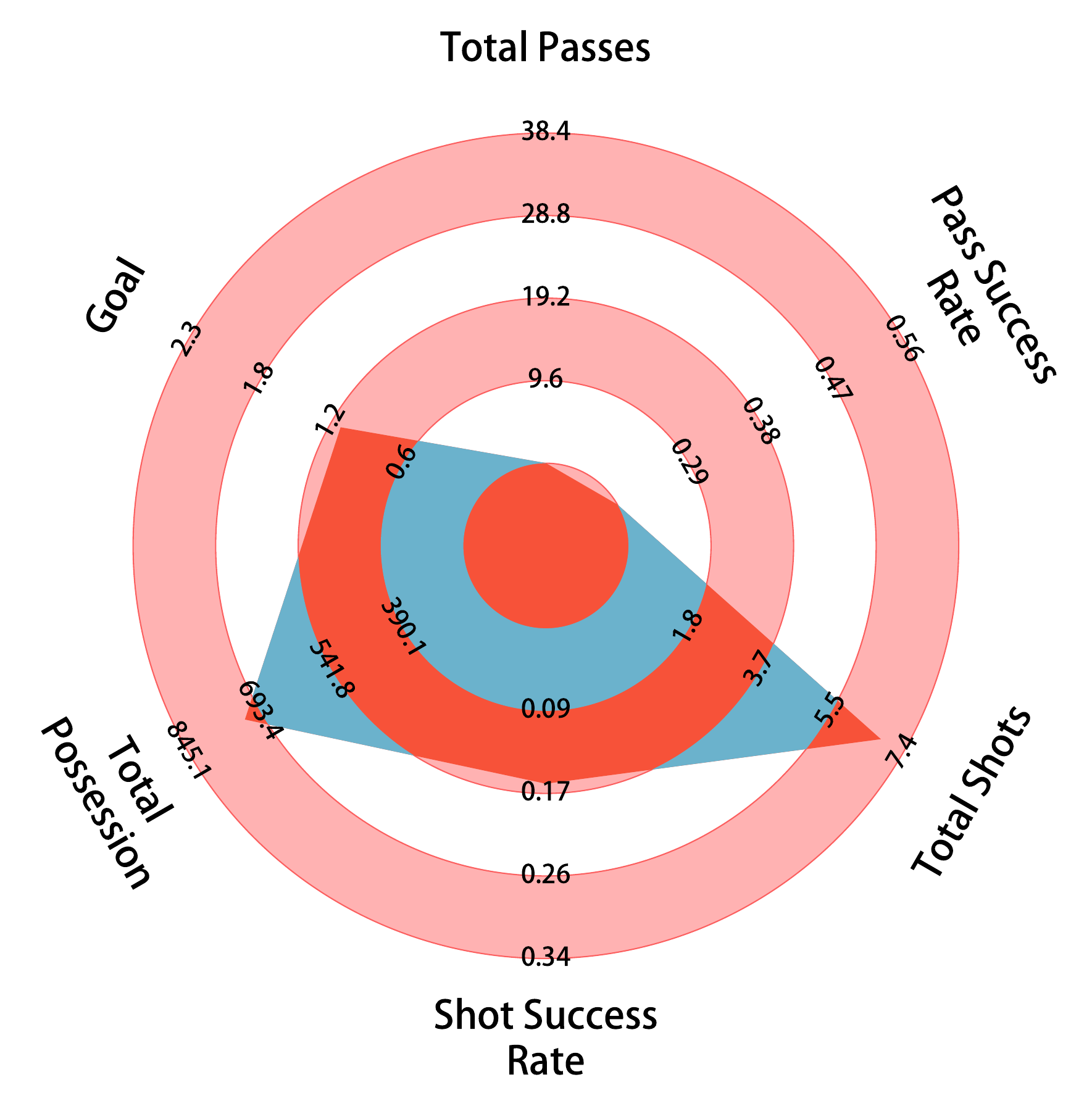}
        \label{fig:formation}
    }
    \caption{Comparison of the policy styles trained using different VLM-based generic potential functions. Their propensity are attack, defend, dribble, and formation, respectively. The detailed instructions provided to VLM can be found in the Appendix~\ref{app: prompt}.}
    \label{fig:single-styles}
\end{figure}

\subsection{Ablation on Adaptive Skill Selection}
To validate the necessity of the adaptive skill selection module, we compare the policies guided by V-GEPF and a fixed VLM-based potential function using radar charts (Figure~\ref{fig:single-styles}). The charts show that the policies guided by a fixed VLM-based potential function exhibit strong stylistic tendencies, with the policy style aligning well with corresponding instruction. For example, an instruction favoring offence leads to policies with high numbers of shots and shot success rates, while a instruction favoring dribbling results in policies with high pass counts and pass success rates, as this effectively prevents the ball from being intercepted. However, football as a long-horizon complex task, requires a more balanced and adaptive policy. As shown in Figure~\ref{fig:radar-vgepor}, the adaptive skill selection module selects the suitable VLM-based potential function at different training stages, resulting in a more balanced policy style compared to not using the module.

\subsection{Visualization Analysis}
We further analyze the policies learned by V-GEPF through visualization. Our method not only improves the win rate, but also learns policies with standard formation and coherent passing, resulting in effects that align with human common sense in football matches.

In the offensive sequence illustrated in Figure~\ref{visualization}, the team executes three distinct passing patterns: transitioning from the backfield to midfield to the forward line, moving from the center to the flanks and back, and switching from long-distance to short-distance passes.

Specifically, in the first frame, the goalkeeper A initiates a long pass to transfer the ball to right back B. In the second frame, as right back B controls the ball, center back E and right midfielder F move closer to provide support. However, the space created by defensive midfielder C's forward run poses a greater threat. Thus, B delivers a long ball to the vacant space in the center, targeting C. In the third frame, it is noteworthy that with C advancing into the attack, F positions itself to recover in order to protect against a potential counterattack from the opponents. Center midfielder D prepares to receive the ball, and naturally, C plays a short pass to D. Finally, in the fourth frame, D executes a dribble and takes a shot, resulting in a goal.
 
The policy demonstrates that the V-GEPF encourages agents to exhibit cooperative behaviors that align with human cognition. Firstly, agents can establish positioning that adheres to the required formation lines. Secondly, during the passing process, agents are capable of making appropriate supporting runs (such as C, E, and F) which provide the receiver with more options, as well as protective runs (such as F) which prevent counterattacks in the event of an unsuccessful offensive play. Furthermore, agents can actively seek to pass into areas that pose a greater offensive threat rather than simply dribbling the ball individually.

\section{Conclusion}
This paper investigates how to align MARL policies with human common sense in complex, long-horizon tasks. While reward shaping methods have proven effective in guiding policies, modeling common sense as a reward remains a challenge. To address this issue, we propose a vision-based reward shaping method called V-GEPF. V-GEPF employs a hierarchical framework that leverages vision information effectively. At the bottom layer, a VLM is used as a generic potential function to efficiently give real-time guidance. At the top layer, a vLLM reflects on the current policy and adjusts the subsequent guidance accordingly. Through experiments, we demonstrate the effectiveness of our proposed method compared to SOTA methods in both improving win rates and aligning with human common sense.

\bigskip

\bibliography{aaai25}

\section{Reproducibility Checklist}

This paper:
\begin{itemize}
    \item  Includes a conceptual outline and/or pseudocode description of AI methods introduced. (yes)
    \item Clearly delineates statements that are opinions, hypothesis, and speculation from objective facts and results. (yes)
     \item Provides well marked pedagogical references for less-familiare readers to gain background necessary to replicate the paper. (yes)
\end{itemize}
\noindent
Does this paper make theoretical contributions? (yes)

\noindent
If yes, please complete the list below.
\begin{itemize}
  \item All assumptions and restrictions are stated clearly and formally. (yes)
  \item All novel claims are stated formally (e.g., in theorem statements). (yes)
  \item Proofs of all novel claims are included. (yes)
  \item Proof sketches or intuitions are given for complex and/or novel results. (yes)
  \item Appropriate citations to theoretical tools used are given. (yes)
  \item All theoretical claims are demonstrated empirically to hold. (yes)
   \item All experimental code used to eliminate or disprove claims is included. (yes)
\end{itemize}

\noindent
Does this paper rely on one or more datasets? (yes)

\noindent
If yes, please complete the list below.
\begin{itemize}
  \item A motivation is given for why the experiments are conducted on the selected datasets. (yes)
\item All novel datasets introduced in this paper are included in a data appendix. (yes)
\item All novel datasets introduced in this paper will be made publicly available upon publication of the paper with a license that allows free usage for research purposes. (yes)
\item All datasets drawn from the existing literature (potentially including authors’ own previously published work) are accompanied by appropriate citations. (yes)
\item All datasets drawn from the existing literature (potentially including authors’ own previously published work) are publicly available. (yes)
\item All datasets that are not publicly available are described in detail, with explanation why publicly available alternatives are not scientifically satisficing. (yes)

\end{itemize}

\noindent
Does this paper include computational experiments? (yes)

\noindent
If yes, please complete the list below.

\begin{itemize}
  \item Any code required for pre-processing data is included in the appendix. (yes).
\item All source code required for conducting and analyzing the experiments is included in a code appendix. (yes)
\item All source code required for conducting and analyzing the experiments will be made publicly available upon publication of the paper with a license that allows free usage for research purposes. (yes)
\item All source code implementing new methods have comments detailing the implementation, with references to the paper where each step comes from (yes)
\item If an algorithm depends on randomness, then the method used for setting seeds is described in a way sufficient to allow replication of results. (yes)
\item This paper specifies the computing infrastructure used for running experiments (hardware and software), including GPU/CPU models; amount of memory; operating system; names and versions of relevant software libraries and frameworks. (yes)
\item This paper formally describes evaluation metrics used and explains the motivation for choosing these metrics. (yes)
\item This paper states the number of algorithm runs used to compute each reported result. (yes)
\item Analysis of experiments goes beyond single-dimensional summaries of performance (e.g., average; median) to include measures of variation, confidence, or other distributional information. (yes)
\item The significance of any improvement or decrease in performance is judged using appropriate statistical tests (e.g., Wilcoxon signed-rank). (yes)
\item This paper lists all final (hyper-)parameters used for each model/algorithm in the paper’s experiments. (yes)
\item This paper states the number and range of values tried per (hyper-) parameter during development of the paper, along with the criterion used for selecting the final parameter setting. (yes)
  
\end{itemize}

\newpage
\onecolumn
\appendix
\section{Algorithm Details}
\subsection{Algorithm of V-GEPF}\label{app: algorithm}  
V-GEPF can be combined with various multi-agent reinforcement learning algorithms, such as MAPPO and HAPPO. Algorithm~\ref{alg:algorithm1} demonstrates the training process of V-GEPF combined with MAPPO to achieve vision-based multi-agent policy alignment.

\begin{algorithm}[ht]
	\caption{The training algorithm of V-GEPF}
	\label{alg:algorithm1}
		\begin{algorithmic}[1] 
		\STATE \textbf{Input:} batch $D$, number of agents $n$, steps per episode $T$, episodes $K$; skill selection cycle $C$;
		\STATE \textbf{Initialize:} Actor $\pi_{\varrho}$; critic $V_{\Theta}$; replay buffer $\mathcal{B}$; VLM-based generic potential function $\phi$;
        
		\FOR{$k$ =1 to $K$}
			\FOR{$t$ =0 to $T$}
    			\STATE Collect the global state $s_t$;
                    \STATE Visualize $s_t$ as $s_t^G$;
    			\FOR{$i$ =1 to $n$}
        			\STATE Collect local observation $o_t^i$;
        			\STATE Select action $a_t^i$ according to the actor $\pi_{\rho}^i$;
                \ENDFOR
                \STATE Execute the joint action $a_t$ and collect the next joint observation $o_{t+1}$, next state $s_{t+1}$, next state's visualization $s_{t+1}^G$, and reward $r_{t+1}$;
                \STATE  Calculate the potential-based reward $F(s_t^G,s_{t+1}^G|l)=\gamma \phi(s_{t+1}^G|l)-\phi(s_t^G|l)$;
                \STATE Calculate the total reward: $R_{t+1}=r_{t+1}+F(s_t^G,s_{t+1}^G|l)$;
                \STATE Calculate TD target $y_t^i$ and advantage function $A_t^i$;
                \STATE Store $(o_{t}, a_{t},s_{t},s_t^G, R_{t+1},o_{t+1},s_{t+1}, s_{t+1}^G, y_t^i, A_t^i)$ into the buffer $\mathcal{B}$;
                \IF{$t \mod C = 0$}  
                    \STATE Update VLM-based generic potential function $\phi$;
                \ENDIF
            \ENDFOR
            \STATE Sample a random mini-batch from buffer $\mathcal{B}$;
            \STATE Calculate loss function of the critic $L_{V} (\Theta)$ and the actor $L_\pi(\varrho)$;
            \STATE Using the gradient descent algorithm to minimize the loss function and update the network parameters $\varrho$, $\Theta$.
		\ENDFOR
	\end{algorithmic}
\end{algorithm}

Specifically, the loss function of the critic can be described as:
\begin{equation}
\label{critic_l}
    L_{V_i}(\Theta) = \mathbb{E}\left[(y_{t}^i - V_{\Theta}^i(o_{t}^i))^2\right],
\end{equation}
\noindent
where $y_t^i$ is the TD target of the critic for agent $i$ at step $t$:
\begin{equation}
\label{critic_y}
y_{t}^i = R_t  + \gamma V_{\Theta}^i(o_{t+1}^i).
\end{equation}

The loss function of the actor can be described as:
\begin{equation}
\label{poicy_loss}
L_{\pi_{i}}(\varrho) = \mathbb{E}\left[\min\left(\frac{\pi_\varrho^{i}(\cdot|o_{t}^i)}{\pi_{\varrho_{\text{old}}}^i(\cdot|o_{t}^i)}, \text{clip}\left(\frac{\pi_\varrho^{i}(\cdot|o_{t}^i)}{\pi_{\varrho_{\text{old}}}^i(\cdot|o_{t}^i)}, 1 - \epsilon, 1 + \epsilon \right)\right) A_t^i(o_{t}^i, a_{t}^i)\right],
\end{equation}
\noindent
where $\pi_{\varrho_{\text{old}}}^i(\cdot|o_{t}^i)$ is the actor before the parameter update, $\epsilon$ and $\lambda$ are two hyperparameters. $A_t^i(o_{t}^i, a_{t}^i)$ is an advantage function estimated using the Generalized Advantage Estimator (GAE) \cite{schulman2015high}:

\begin{equation}
\label{gae}
A_t^i(o_{t}^i, a_{t}^i) = \sum_{l=1}^{\infty} (\gamma \lambda)^{l} 
\left( R_t + \gamma V_{\Theta}^i(o_{t+l+1}^i) - V_{\Theta}^i(o_{t+l}^i) \right)
\end{equation}

\subsection{Algorithm Proof}\label{app:proof}
To demonstrate that incorporating generic potential functions as potential-based rewards do not alter the optimal response in multi-agent reinforcement learning (MARL), we first prove the preservation of policy invariance in single-agent problems. Then we prove the maintenance of consistent Nash equilibrium in multi-agent problems.

\begin{proof}
Take an arbitrary agent from all agents as an example. Given that introducing potential-based reward shaping only alters the reward function element in the POMDP, the two agent $L$ and $L'$ before and after the reward shaping have the same sequence of states and actions $\overline{s}=\{s_0,a_0,...,s_N\}$ in an episodic environment. The value function of agent $L$ can be defined as:
\begin{equation}
\label{value_return}
    V=\mathbb{E}_{\pi} \left [ U_L(\overline{s}) \right]=\mathbb{E}_{\pi} \left [\sum_{t=0}^{N-1} \gamma^t r_{t}  \right].
\end{equation}
\noindent
where $r_t$ represents the environmental reward of agent $L$.

When leveraging the generic potential-based reward $F(s,s'|l)=\gamma \phi(s'|l)-\phi(s|l)$ to reshape the environmental reward, a different return $U_{L'}$ experiencing the same sequence is calculated as:
\begin{equation}
\label{poten_u}
\begin{split}
    U_{L'}(\overline{s}) &= \sum_{t=0}^{N-1} \gamma^t \left( r_t + F(s,s'|l) \right)\\
    &= \sum_{t=0}^{N-1} \gamma^t \left( r_t + \gamma\phi(s'|l)-\phi(s|l) \right)\\
    &= \underbrace{\sum_{t=0}^{N-1} \gamma^t r_t}_{U_L(\overline{s})} + \sum_{t=0}^{N-1} \gamma^{t+1} \phi(s'|l) - \sum_{t=0}^{N-1} \gamma^t \phi(s|l)\\
    &= U_L(\overline{s}) + \sum_{t=1}^{N-1} \gamma^t \phi(s|l)+\gamma^{N} \phi(s_N|l) \\
    &- \phi(s_0|l) - \sum_{t=1}^{N-1} \gamma^t \phi(s|l)\\
    &= U_L(\overline{s}) + \gamma^{N} \phi(s_N|l) - \phi(s_0|l).
\end{split}
\end{equation}

It is worth noting that the potential function at the step $N$ satisfies the following conditions $\phi(s_N|l)=0$, as outlined in the paper. Therefore, the relationship between cumulative discount returns can be simplified as: 
\begin{equation}
\label{return_relation}
    U_{L'}(\overline{s})= U_L(\overline{s}) - \phi(s_0|l).
\end{equation}

Given that $\phi(s_0|l)$ is the initial state which is definitely independent of the policies $\pi$, the relationship between the value functions is as follows based on Equation \ref{value_return} and \ref{return_relation}:
\begin{equation}
\label{value_relation}
    V_{L'}(s)=V_L(s)-\phi(s_0|l).
\end{equation}

Agent $L$ will update its state-value function by:
\begin{equation}
\label{value_update_1}
\begin{split}
    V_L(s) \leftarrow V_L(s) + \alpha \underbrace{\left( r_t + \gamma V_L(s') - V_L(s) \right)}_{\delta V_L(s)},
\end{split}
\end{equation}
\noindent
where $\Delta V_L(s)=\alpha \delta V_L(s)$ represents the amount by which the state-value functions are updated. 

Thus, the current values of agent $L$ can be represented as:
\begin{equation}
    V_{L}(s) = V_{0}(s) + \Delta V_{L}(s),
\end{equation}
\noindent
where $V_{0}(s)$ is the initial value. 

After incorporating potential-based rewards, the value function is updated by:
\begin{equation}
\label{value_update_2}
\begin{split}
    V_{L'}(s) \leftarrow V_{L'}(s) + \alpha \underbrace{\left( r_t + F(s,s'|l) + \gamma V_{L'}(s') - V_{L'}(s) \right)}_{\delta V_{L'}(s)}.
\end{split}
\end{equation}

Similarly, the current value of agent $L^{\prime}$ is:
\begin{equation}
    V_{L'}(s) = V_{0}(s) - \phi(s|l) + \Delta V_{L'}(s).
\end{equation}

At the beginning, both $\Delta V_L(s)$ and $\Delta V_{L'}(s)$ equal zero. When the agent $L$ starts state transition, it will update its values by:
\begin{equation}
\label{value_update_3}
\begin{split}
    \delta V_L(s) = &r_t + \gamma V_L(s') - V_L(s)\\
    = &r_t + \gamma (V_0(s') + \Delta V_L(s'))- V_0(s) - \Delta V_L(s).
\end{split}
\end{equation}

The agent $L'$ will update its values by:
\begin{equation}
\label{value_update_4}
\begin{split}
    \delta V_{L'}(s) =& r_t +\gamma\phi(s'|l)-\phi(s|l)+ \gamma V_{L'}(s') - V_{L'}(s)\\
    = &r_t + \gamma\phi(s'|l)-\phi(s|l)+\gamma (V_0(s')-\phi(s'|l)+ \Delta V_{L'}(s'))\\
    &- V_0(s) +\phi(s|l)- \Delta V_{L'}(s)\\
    =&r_t +\gamma (V_0(s') + \Delta V_{L}(s'))- V_0(s) - \Delta V_{L}(s)\\
    =&\delta V_{L}(s).
\end{split}
\end{equation}

Sequentially and recursively, the critic of agents $L$ and $L'$ are updated with the same values. Thus, we have the equation that:
\begin{equation}
\label{value_update_5}
\begin{split}
    \pi^*=argmax_{\pi \in \Pi}E_{s_0 \sim \rho}[V_L(s_0)]=argmax_{\pi \in \Pi}E_{s_0 \sim \rho}[V_{L'}(s_0)],
\end{split}
\end{equation}
\noindent
where $\rho$ is the distribution of $s_0$.

Therefore, the generic potential-based rewards do not altering the optimal policy.
\end{proof}

Subsequently, we need to prove that potential-based reward shaping does not change the Nash equilibrium in a MARL environment.
\begin{proof}
    In MARL, a Nash equilibrium can be Formally defined as:
    \begin{equation}
        \label{nash}
        R_i(\pi^{\text{NE}}_i \cup \pi^{\text{NE}}_{-i}) \geq R_i(\pi_i \cup \pi^{\text{NE}}_{-i}), \forall i \in 1, \ldots, n, \pi_i \in \Pi_i
    \end{equation}
    \noindent
    where $n$ is the number of agents, $\Pi_i$ is the agent $i$'s set of possible policies, $R_i$ is the agent $i$'s reward function, $\pi^{NE}_i$ and $\pi^{NE}_{-i}$ represent a specific policy of agent $i$ and the joint policy of all other agents except agent $i$. When the conditions of Eq. \ref{nash} are met, the joint policy where each agent follows its optimal policy $\pi^{NE}_i$ constitutes a Nash equilibrium.

    For agent $i$, we consider that its joint policies $\Pi_i^{NE}$ include its each possible policy combined with $\pi_{-i}^{NE}$, which can be described as follows:
    \begin{equation}
        \label{pi}
        \pi_i \cup \pi_{-i}^{NE}, \forall \pi_i \in \Pi_i.
    \end{equation}
    \noindent
    Each joint polciy $\Pi_i^{NE}$ can generate a fixed experience sequence as: 
    \begin{equation}
        \label{sequence}
        \overline{s}=\{s_t^i,a_t^i,r_t^i\}| t=0,1,...N-1,i=0,1, ...,n.
    \end{equation}
    \noindent
    Based on the Eq. \ref{poten_u}, any policy that satisfies Eq. \ref{nash} will continue to satisfy it, and the following conclusion can be derived:
    \begin{equation}
        \label{nash_last}
        \begin{split}
            &(R_i(\pi^{NE}_i \cup \pi^{{NE}}_{-i}) \geq R_i(\pi_i \cup \pi^{\text{NE}}_{-i})) \\
            \leftrightarrow &(R_{i,\phi}(\pi^{{NE}}_i \cup \pi^{{NE}}_{-i}) \geq R_{i,\phi}(\pi_i \cup \pi^{{NE}}_{-i})) , \forall \pi_i \in \Pi_i.
        \end{split}
    \end{equation}

    Since reward shaping only modifies the specific agent's reward function and does not affect other agents, those unaffected agents will continue to maintain their policies as part of the Nash equilibrium. Therefore, regardless of the number of agents in the multi-agent system that implement or do not implement potential-based reward shaping, the nash equilibrium will remain stable and unchanged.
\end{proof}

\newpage
\section{Implementation Details}
\subsection{Visualization of States}\label{app: CLIPimage}

Knowledge of the soccer domain is aggregated to the visualisation of states for more image representation. 
The pitch is first visualised with mplsoccer\footnote{\url{https://github.com/andrewRowlinson/mplsoccer}}. In detail, the players and ball are represented as circles with blue, red and black colors, corresponding to the offensive team, the defensive team and the ball, respectively.
To further indicate the team formation and the contact between players, we visualize a convex hull and formation lines. The convex hull is the coverage area for all players except the goalkeeper, and reflects the ability to cover the pitch between formations. The formation lines are sequential lines of player combinations with the same responsibilities and reflects the discipline and stability within formations.

\begin{figure}[h]
    \centering
    \includegraphics[width=0.5\linewidth]{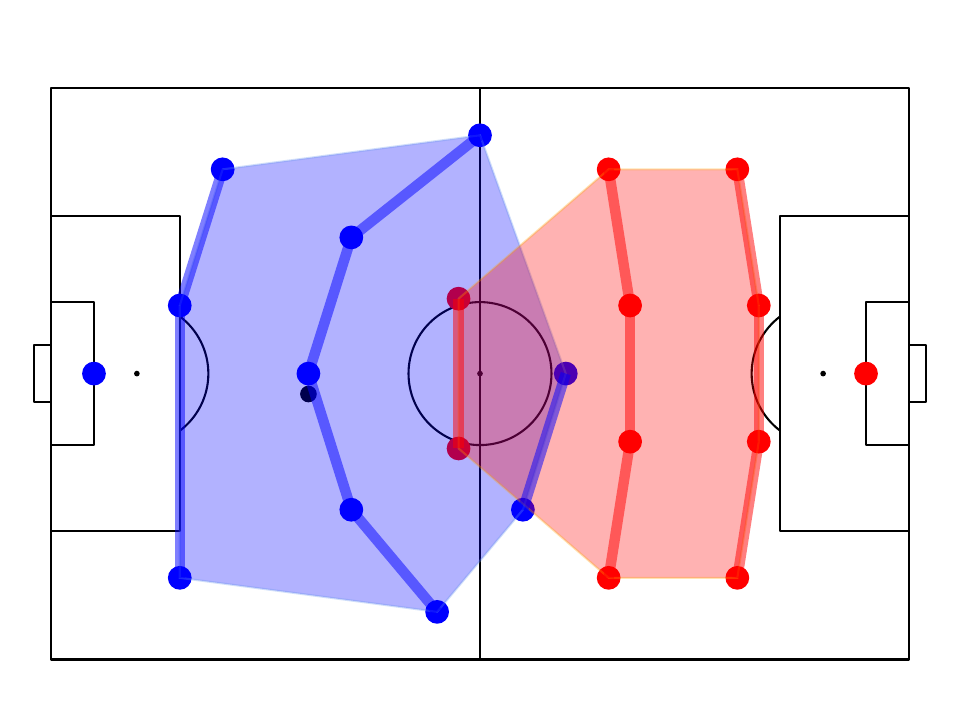}
    \caption{Example of VLM's image input}
    \label{fig:example_visual}
\end{figure}

In the Figure \ref{fig:example_visual}, the offensive team lines up in a 3-5-2 with the defensive midfielder (DM) possessing the ball, and the defensive team lines up in a 4-4-2. The blue convex hull is larger than the red corresponding to more aggression.

\subsection{Prompt Details of VLM}\label{app: prompt}
Based on the domain knowledge of football and the current capabilities of VLMs, we have categorized the the potential function pool into four types. These four types aim to comprehensively cover the various aspects where potential functions might guide the algorithm.

\begin{itemize}
    \item Macro level: this category of potential functions quantifies the macro-level situation. The prompt include: "The blue team has a bigger advantage than the red team."
    \item Micro level, state-based: these potential functions focus on analyzing microscopic state information, quantifying local situations. The prompt include: "The ball is close to the opponent's goal."
    \item Micro level, agent-based: these potential functions focus on the behaviors and interactions of agents, quantifying the proficiency of specific agent skills. The prompts include: ["The blue team is performing a coordinated attack.", "The blue team is trying to defend when the ball is close to their goal.", "Three blue formation lines are parallel and have proper spacing.", "The black ball is well dribbled by the blue team."]
    \item Micro level, goal-based: these potential functions focus on specific objectives (e.g., the number of shots, the number of passes). Traditional rule-based potential functions often fall into this category. By using VLM-based potential functions combined with appropriate visualizations, we can cover the range of capabilities of rule-based methods. The prompt include: "the blue team is passing"
\end{itemize}

We believe this classification could also inspire other tasks beyond football. It is worth noting that all potential functions use the same VLM, and we switch prompts to represent different potential functions.

\subsection{Dialogue Example of vLLM}\label{app: dialogue}
The full prompt contains a video part and text part. The text part includes instructions related to the initial goal, records of the training process, information about the potential function pool, and the record of the last potential function as feedback for reflection. In our experiments, vLLM is called every 50 epochs to select the next appropriate skill. The records in the text prompt include the average values of corresponding variables in each epoch for the 50 epochs.

\begin{center}
    \begin{tcolorbox}[colback=gray!10,
                      colframe=black,
                      width=15cm,
                      arc=1mm, auto outer arc,
                      boxrule=0.5pt,
                      title={Full text prompt of vLLM}
                     ]

Analyse this video.
We want to guide the blue team to master human-like football skills. Here are some useful information:

Here are some records:

total shot: 
[0.0, 0.0, 0.0, 0.0, 0.0, 0.0, 0.0, 0.0, 0.0, 0.03, 0.03, 0.0, 0.0, 0.0, 0.0, 0.0, 0.0, 0.06, 0.03, 0.06, 0.0, 0.0, 0.03, 0.09, 0.06, 0.09, 0.03, 0.06, 0.16, 0.22, 0.09, 0.09, 0.28, 0.12, 0.22, 0.19, 0.25, 0.47, 0.59, 0.28, 0.53, 0.56, 0.41, 0.78, 0.5, 1.03, 1.12, 1.25, 1.25, 1.0]

win rate: 
[0.0, 0.0, 0.0, 0.0, 0.0, 0.0, 0.0, 0.0, 0.0, 0.0, 0.0, 0.0, 0.0, 0.0, 0.0, 0.0, 0.0, 0.0, 0.0, 0.0, 0.0, 0.0, 0.0, 0.0, 0.0, 0.0, 0.0, 0.0, 0.0, 0.0, 0.0, 0.0, 0.0, 0.0, 0.03, 0.0, 0.0, 0.0, 0.0, 0.03, 0.0, 0.03, 0.0, 0.0, 0.0, 0.0, 0.0, 0.0, 0.03, 0.0]

Please carefully analyze the record and current video, then provide advice for the next skill.

You should choose one from [has advantage, ball location, correct formation, encourage dribbling, encourage attack, encourage defense, encourage passing]. Here are their descriptions:

has advantage: the blue team has bigger advantage than the red team.

ball location: the ball is close to the opponent’s goal.

correct formation: three blue formation lines are parallel and have proper spacing.

encourage dribbling: the black ball is well dribbled by the blue team.

encourage attack: the blue team is performing a coordinated attack.

encourage defense: the blue team is trying to defend when the ball is close to their goal.

encourage passing: the blue team is passing.

The last skill is encourage attack. Its reward record is [-0.0, 0.01, -0.02, 0.0, 0.02, 0.02, -0.0, 0.03, 0.02, 0.04, 0.02, -0.0, 0.04, 0.03, 0.03, 0.05, 0.02, 0.06, 0.06, 0.07, 0.1, 0.15, 0.22, 0.22, 0.2, 0.23, 0.22, 0.24, 0.24, 0.2, 0.24, 0.24, 0.24, 0.23, 0.24, 0.26, 0.23, 0.24, 0.24, 0.25, 0.23, 0.25, 0.26, 0.23, 0.25, 0.25, 0.25, 0.23, 0.23]. What is the next skill to master? Try not repeating the last skill.

Output Format:

Analysis: ...

Next Skill: ...

    \end{tcolorbox}
\end{center}

\begin{center}
    \begin{tcolorbox}[colback=gray!10,
                      colframe=black,
                      width=15cm,
                      arc=1mm, auto outer arc,
                      boxrule=0.5pt,
                      title={vLLM's response}
                     ]
Analysis: To encourage the blue team to master human-like football skills, focusing on coordination and strategy is essential. The record shows that the last skill, which was likely "encourage attack," has been rewarded with high scores in recent actions. Given this context, the next skill for the blue team to master should be "correct formation." This would help the team develop a structured approach to the game, ensuring that they are well-positioned to receive passes and create scoring opportunities. A strong formation can also aid in defense, allowing the team to maintain a solid shape when under pressure from the opposing team.

Next Skill: Correct formation
    \end{tcolorbox}
\end{center}

\newpage
\section{Experimental Details}
\subsection{Experimental Environments and Scenarios}\label{app:scenarios}
\textbf{Google Research Football Environment.} Google Research Football (GRF) \cite{kurach2020google} is a benchmark reinforcement learning environment, in which the agents are trained to play football and defeat the opponents controlled by built-in script. GRF provides three types of state representations to each agent: pixels, super mini map and raw observation of 115-dimensional vector. The paper adopts the 115-dimensional vector as the observation, which includes many complicated and comprehensive information of the match, such as the position and the speed of each player and the ball. Based on the state representations, there are 19 actions to be chosen for each agent, including move actions towards 8 directions, and 11 different kicking techniques. GRF provides a set of football full scenarios for efficient training. These scenarios have distinct initial situations, with various players and opponents positioned at different starting locations, and the opponents have adjustable difficulty settings.

\begin{figure}[h]
  \centering
    \subfigure[]{              
        \includegraphics[width=5cm]{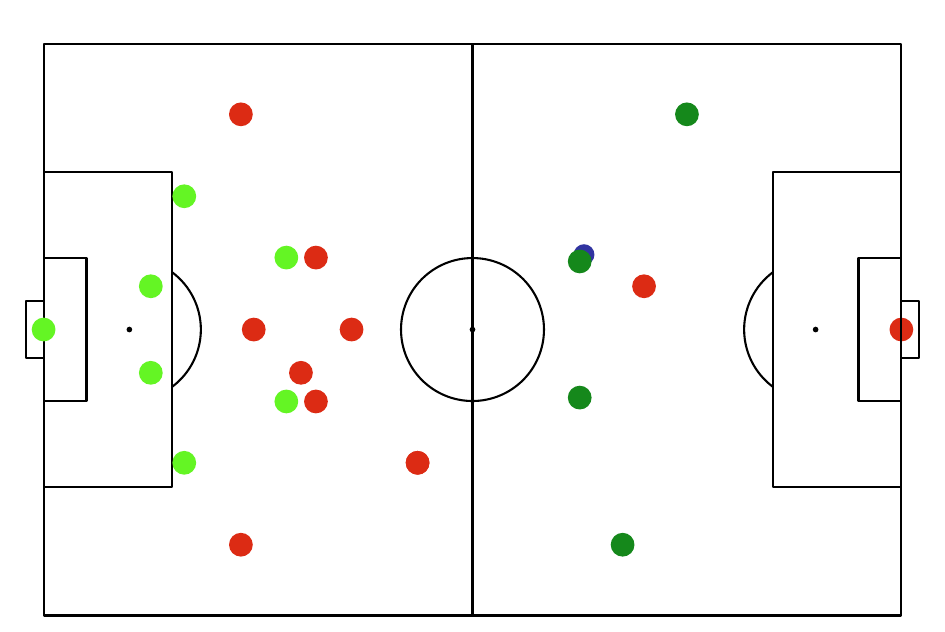}
        \label{fig:easy}}
    \subfigure[]{
        \includegraphics[width=5cm]{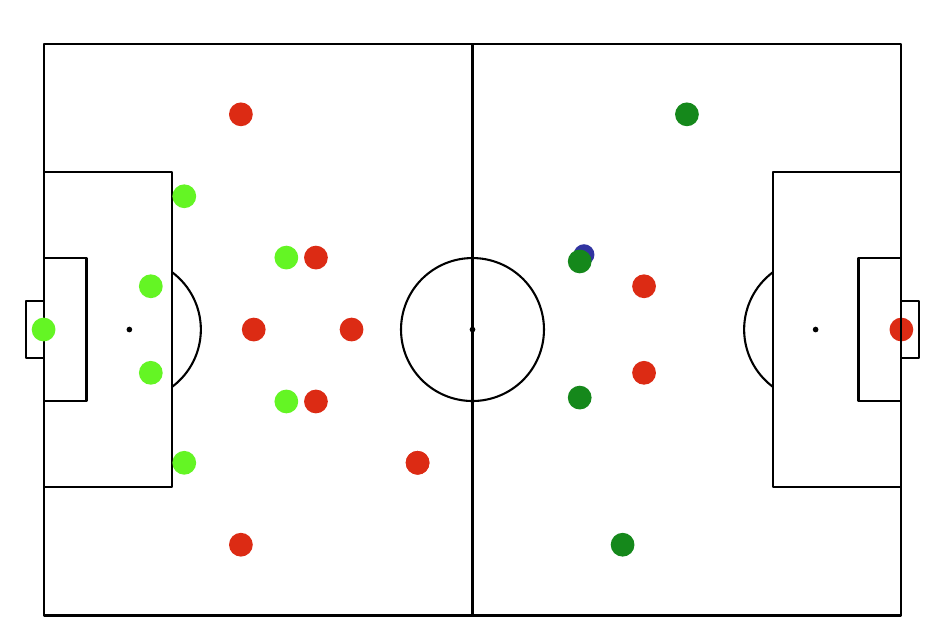}
        \label{fig:hard}}
    \subfigure[]{
        \includegraphics[width=5cm]{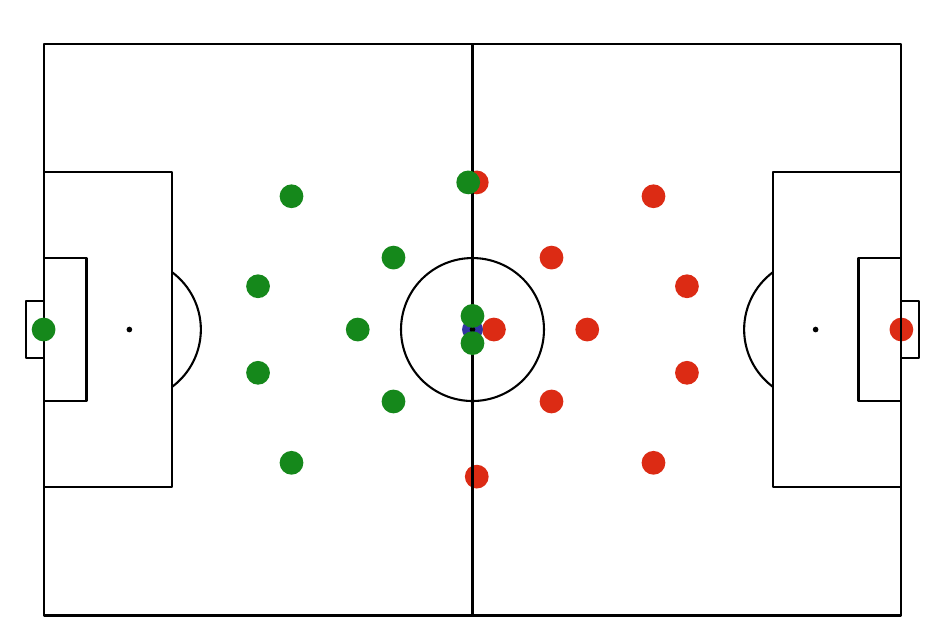}
        \label{fig:full}}
    \caption{Visualization of the initial position of each agent, opponent and the ball in four GRF scenarios. The deep green dots represent the agents, the light green dots represent the other teammates of the agents, the red dots represent the opponents, and the blue dot denotes the ball. (a) Academy counterattack\_easy, (b) Academy counterattack\_hard, (c) 11{\_}vs{\_}11.}
    \label{fig:scenarios}
\end{figure}

\textbf{GRF Scenarios.} The GRF environment simulates real football matches, which serves as benchmarks for football game AI and enables researchers to assess the performance of their algorithms. Four different simulation scenarios are chosen in this paper, i.e., {\bf{\textit{academy counterattack{\_}easy, academy counterattack{\_}hard, academy 11{\_}vs{\_}11{\_}easy,}}} and {\bf{\textit{academy 11{\_}vs{\_}11{\_}hard.}}} In these scenarios, the left team is controlled by MARL algorithm. In different scenarios, the agents, the opponents, and the ball have different initial positions which are shown in Figure~\ref{fig:scenarios}. \textit{academy 11{\_}vs{\_}11} and \textit{academy counterattack} are both full-field scenarios. In \textit{academy counterattack}, only the four players in the front (marked in deep green) are controlled by MARL algorithm and others are controlled by built-in script, while all 11 players are controlled by MARL algorithm in \textit{academy 11{\_}vs{\_}11}. Specifically, in \textit{academy counterattack} task, the agents adopt roles of left midfield, central midfield, right midfield, and central front respectively. \textit{11{\_}vs{\_}11{\_}easy} and \textit{11{\_}vs{\_}11{\_}easy} task have opponents' difficulty of $0.05$ and $0.95$ respectively.

In GRF, all agents must cooperate effectively to organize offenses, counter the opponents' policies, and score goals. A $+1$ reward when the left team scores a goal. An episode will terminate when one of the following four conditions occurs: (1) a goal is scored, (2) the ball goes out of bounds, (3) the number of steps reaches the set limit. The condition for success is that the left team (i.e., the team with the agents) scores more goals.

\subsection{Hyperparameters}\label{app: hyperparameters}

\begin{table}[h!] 
\caption{Hypreparameters of algorithms used in experiments.}
\begin{center}
\begin{tabular}{c|c|c}
\hline
\textbf{Hyperparameter} & \textbf{Value} & \textbf{Description} \\ \hline
Network & MLP & \multirow{5}{*}{Actor Network} \\ 
Dimensions of hidden layers & $[256,128,64]$ & \\ 
Activation & ReLU & \\
Optimizer & Adam & \\
Learning rate & $5e-4$ & \\ \hline
Network & MLP & \multirow{5}{*}{Critic Network} \\ 
Dimensions of hidden layers & $[256,128,64]$ & \\ 
Activation & ReLU & \\
Optimizer & Adam & \\
Learning rate & $5e-4$ & \\ \hline
Rollout length & $3001$ & \multirow{7}{*}{RL Parameters} \\ 
Number of workers & $32$ & \\ 
Batch size & $32$ & \\ 
Discounting factor & $0.995$ & \\
PPO epoch & 10 & \\
GAE lambda & 0.95 & \\
PPO clip & 0.2 & \\ \hline
Reward coefficient $\rho$ & $0.5$ & Exclusive to V-GEPF  \\ \hline
Reward coefficient $\rho$ & $0.02$ & Exclusive to MAPPO-xT \\ \hline

\end{tabular}
\label{tab1}
\end{center}
\end{table}

Hyperparameters are shown in Table~\ref{tab1}. We used the same parameters for V-GEPF, MAPPO-xT, MAPPO, IPPO, and HAPPO, except for algorithm-specific parameters. Our baselines are implemented based on the repository DB-Football\footnote{\url{https://github.com/Shanghai-Digital-Brain-Laboratory/DB-Football}}.

\subsection{Expected Threat (xT)} \label{app: xT}

Expected Threat (xT) is a novel metric to quantify the threat or danger created by a team or player in possession of the ball. Unlike traditional metrics like Expected Goals (xG) which focus solely on shot quality, xT aims to capture the potential for a possession to lead to a goal, even if it doesn't result in a shot.

The key idea behind xT is to model a goal as a Markov chain constructed by a series of passes and a shot. The pitch is divided into a $21 \times 15$ blocks. From any given block, there are passing possibilities to the remaining $21 \times 15 - 1$ blocks. The xT of a particular block is the sum of the probabilities of scoring directly from a shot or passing to another block and eventually scoring. When a goal occurs, the goal value (set to 1) propagates along the Markov chain. Formally, the xT of a zone $(x,y)$ is defined as:

\begin{equation}
\mathrm{xT}_{x, y}=\left(s_{x, y} \times g_{x, y}\right)+\left(m_{x, y} \times \sum_{z=1}^{21} \sum_{w=1}^{15} T_{(x, y) \rightarrow(z, w)} \mathbf{x} \mathrm{T}_{z, w}\right).
\end{equation}

where $s_{x,y}$ is the probability of taking a shot from zone $(x,y)$, $g_{x,y}$ is the probability of scoring from a shot in zone $(x,y)$, $m_{x,y}$ is the probability of passing the ball from zone $(x,y)$, $T_{(x,y)\rightarrow(z,w)}$ is the probability of passing the ball from zone $(x,y)$ to zone $(z,w)$ given that a pass is made.

To compute xT, we start with all zones having an xT of 0 and iteratively update the values using the above formula until convergence, typically in 4-5 iterations. This allows us to break the cyclic dependency and gives xT a natural interpretation - the probability of scoring within the next $n$ actions, where $n$ is the iteration number.

\begin{figure}[h]
    \centering
    \includegraphics[width=0.5\linewidth]{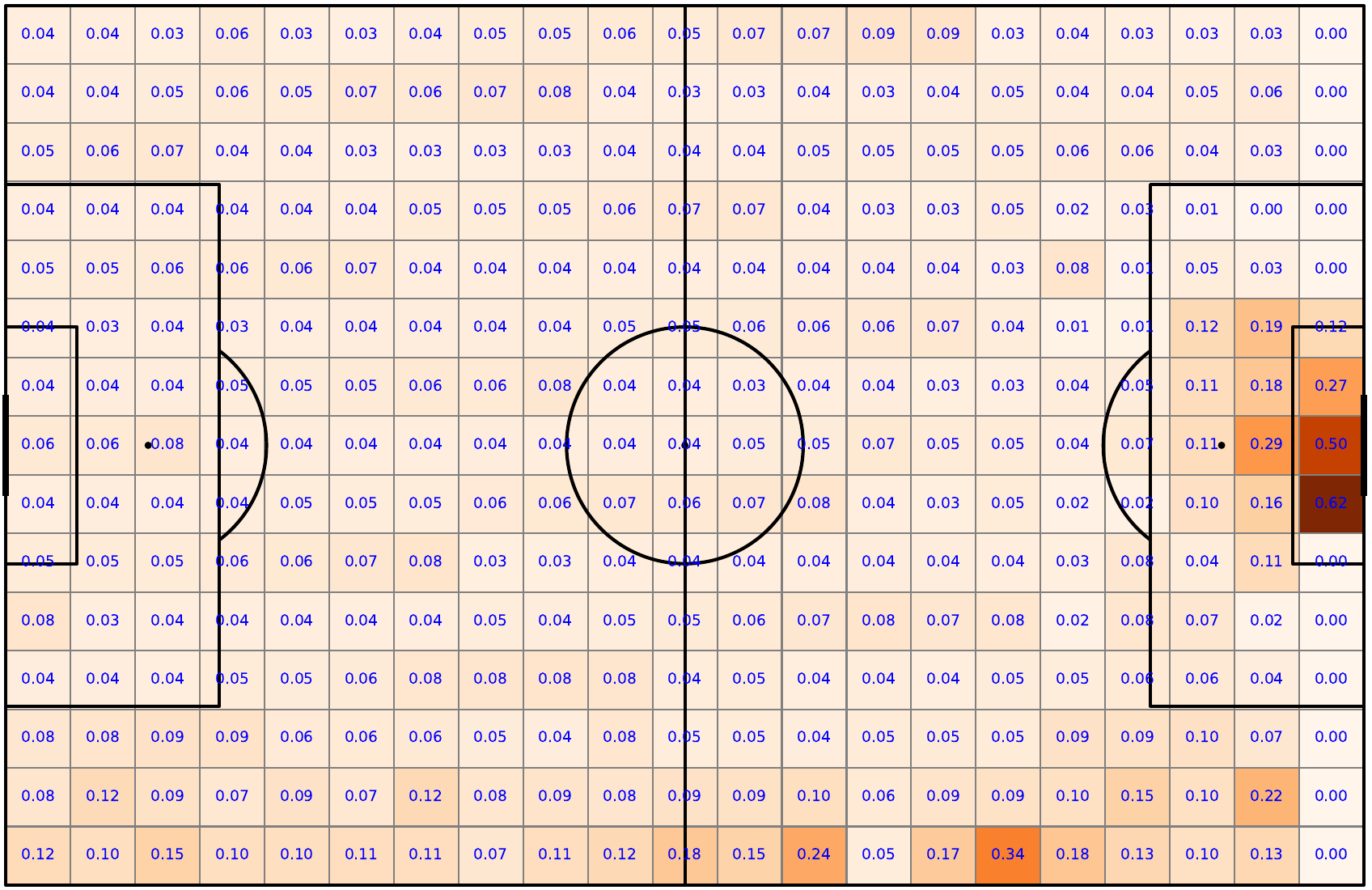}
    \caption{xT values after 5 iterations, calculated using data from the FIFA World Cup 2018.}
    \label{fig:enter-label}
\end{figure}

\subsection{Supplementary Experiments}\label{app: supplementary_experiments}
\subsubsection{Analysis on Coefficient $\rho$}
\begin{figure}[t!]
    \centering
    \includegraphics[width=0.4\linewidth]{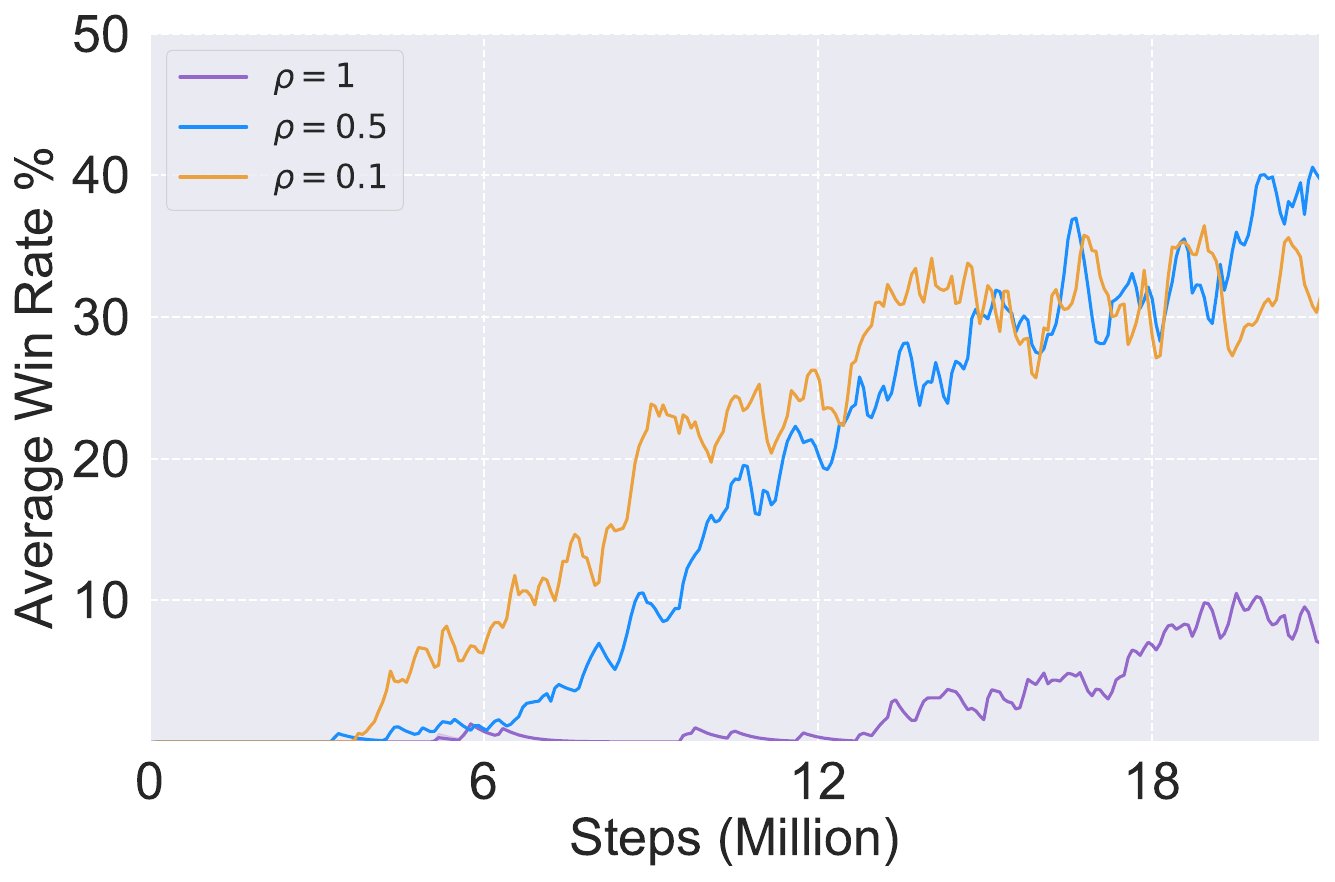}
    \caption{Analysis on coefficient $\rho$ of V-GEPF.}
    \label{fig:ab_rho}
\end{figure}

We conducted a parameter analysis in the GRF 11vs11 hard scenario, searching for the optimal value of the parameter $\rho$ within [1, 0.5, 0.1]. The results are shown in Figure~\ref{fig:ab_rho}. In this scenario, the goal reward is sparse, valued at 1, while the potential function reward is dense, necessitating careful tuning of the coefficient $\rho$. When $\rho$ is set to 1, the potential function reward becomes overly dominant, diminishing the influence of the goal reward and resulting in a low win rate. Conversely, if $\rho$ is set too low, its ability to guide policy learning is weakened. Specifically, when $\rho$ is set to 0.1, the final win rate is lower than when $\rho$ is set to 0.5. Therefore, we chose to set $\rho$ in V-GEPF to 0.5.

\subsubsection{Additional Experiments on \textit{academy counterattack}}
To further evaluate our approach, we conducted additional experiments in a simpler full-field scenario \textit{academy counterattack}. In this scenario, the RL algorithm controls only four out of our eleven players, while the remaining seven players are controlled by a built-in script. This setup significantly reduces the exploration space while still maintaining the dynamics of a full-field match. We evaluate whether V-GEPF remains effective in this scenario. 

In this experiment, we set rollout workers to 80, with a rollout length of 400, and train for 200 epochs, resulting in a total of 6.4 million exploration steps. Due to the shorter rollout length compared to the 11{\_}vs{\_}11 scenario, we set the reward coefficient $\rho=1$. To control for the effects of randomness, all curves were generated using the same random seed. 

\begin{figure}[h]
    \centering
    \subfigure[\textit{academy counterattack} easy]{              
        \includegraphics[width=6cm]{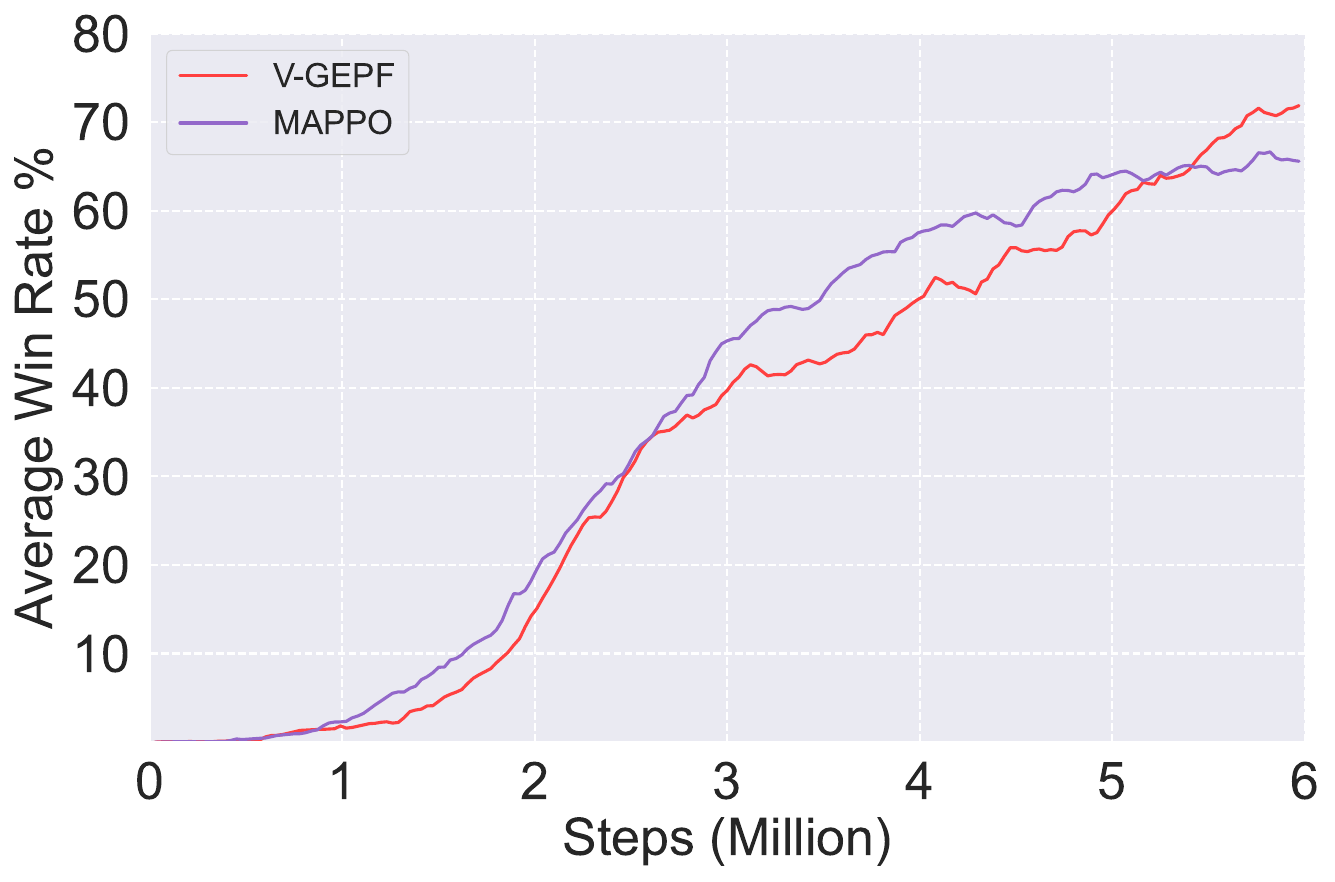}
        \label{add_easy}}
    \subfigure[\textit{academy counterattack} hard]{
        \includegraphics[width=6cm]{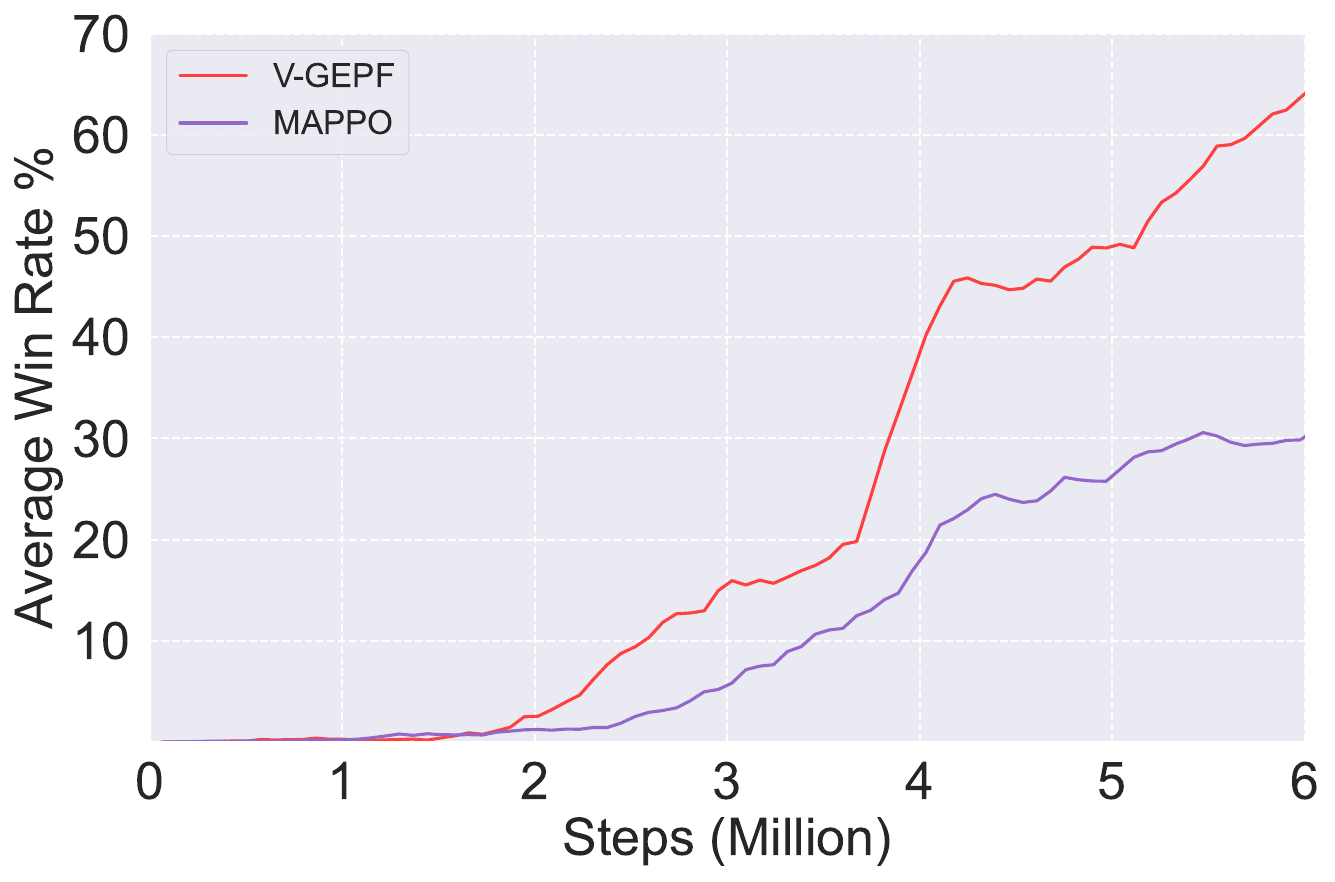}
        \label{add_hard}}
    \caption{Average win rate curves during training in GRF \textit{academy counterattack} scenario.}
    \label{add_exp}
\end{figure}

As shown in the average win rate curve in Figure~\ref{add_exp}, MAPPO quickly converges to a relatively high win rate. However, V-GEPF consistently achieves a higher win rate than MAPPO, both in easy and hard scenarios, demonstrating the effectiveness of our proposed method in \textit{academy counterattack}. It is worth noting that we observe a similar phenomenon as in the 11{\_}vs{\_}11 scenario: the improvement of V-GEPF over MAPPO is more significant in the hard task than in the easy one. This suggests that rewards in the hard task are more sparse, making the guidance from the general potential function more essential.

\end{document}